# A Sparse Model-inspired Deep Thresholding Network for Exponential Signal Reconstruction—Application in Fast Biological Spectroscopy

Zi Wang, Di Guo, Zhangren Tu, Yihui Huang, Yirong Zhou, Jian Wang, Liubin Feng, Donghai Lin, Yongfu You, Tatiana Agback, Vladislav Orekhov, Xiaobo Qu*

*Abstract* — The non-uniform sampling is a powerful approach to enable fast acquisition but requires sophisticated reconstruction algorithms. Faithful reconstruction from partial sampled exponentials is highly expected in general signal processing and many applications. Deep learning has shown astonishing potential in this field but many existing problems, such as lack of robustness and explainability, greatly limit its applications. In this work, by combining merits of the sparse model-based optimization method and data-driven deep learning, we propose a deep learning architecture for spectra reconstruction from undersampled data, called MoDern. It follows the iterative reconstruction in solving a sparse model to build the neural network and we elaborately design a learnable soft-thresholding to adaptively eliminate the spectrum artifacts introduced by undersampling. Extensive results on both synthetic and biological data show that MoDern enables more robust, high-fidelity, and ultra-fast reconstruction than the state-of-the-art methods. Remarkably, MoDern has a small number of network parameters and is trained on solely synthetic data while generalizing well to biological data in various scenarios. Furthermore, we extend it to an open-access and easy-to-use cloud computing platform (XCloud-MoDern), contributing a promising strategy for further development of biological applications.

## I. INTRODUCTION

Fast data acquisition of exponentials is widely used in many signal processing applications: Telecommunication [1, 2], fluorescence microscopy [3], analog-to-digital conversion [4, 5], medical imaging [6], geoscience [7], biological nuclear magnetic resonance (NMR) spectroscopy [8-10], and so on. The non-uniform sampling (NUS) has been a popular technique to enable fast acquisition by reducing the amount of acquired data [11, 12]. The undersampling process of exponential signal $\mathbf{r} \in \mathbb{C}^N$ can be mathematically modeled as

$$\mathbf{y} = \mathbf{U}\mathbf{r} + \boldsymbol{\varepsilon}, \quad (1)$$

where $\mathbf{y} \in \mathbb{C}^M$ is the time-domain signal undersampled by matrix $\mathbf{U} \in \mathbb{R}^{M \times N}$, $\boldsymbol{\varepsilon} \in \mathbb{C}^M$ is the additive noise. The NUS density is defined as $SR = (M/N) \times 100\%$, where $M, N$ are the number of partially sampled and fully sampled data points, respectively. Since the data is undersampled, there are lots of artifacts in the spectrum obtained by Fourier transform of $\mathbf{y}$. Thus, faithful reconstruction from undersampled exponential signals is one of the frontiers and highly significant problem in signal processing [13].

Over the past two decades, many model-based reconstruction methods have been established. They are well theoretically grounded and assume specific signal priors with insights from exponential signals or the corresponding spectra. A range of modern approaches include the maximum entropy [12], spectral lineshape estimation [14], tensor structures [11], compressed sensing (CS) [15-20], and low-rank Hankel reconstruction [8-10, 21, 22]. These methods can reconstruct spectra well, but several practical limitations and conceptual questions remain. Although implemented differently, they are almost iterative and require lengthy computational time to perform reconstruction. Moreover, certain assumptions of signal prior may not be

This work was supported in part by the National Natural Science Foundation of China under grants 6212200447, 61971361, 61871341, and 61811530021, the National Key R&D Program of China under grant 2017YFC0108703, the Xiamen University Nanqiang Outstanding Talents Program, the Swedish Research Council under grant 2015–04614, and the Swedish Foundation for Strategic Research under grant ITM17-0218.

Zi Wang, Zhangren Tu, Yihui Huang, Yirong Zhou, Jian Wang, and Xiaobo Qu* are with the Department of Electronic Science, Biomedical Intelligent Cloud R&D Center, Fujian Provincial Key Laboratory of Plasma and Magnetic Resonance, National Institute for Data Science in Health and Medicine, Xiamen University, China (*Corresponding author, email: quxiaobo@xmu.edu.cn).

Di Guo is with the School of Computer and Information Engineering, Xiamen University of Technology, Xiamen, China.

Liubin Feng and Donghai Lin are with the College of Chemistry and Chemical Engineering, Key Laboratory for Chemical Biology of Fujian Province, High-field NMR Center, Xiamen University, Xiamen, China.

Yongfu You is with the China Mobile Group, Xiamen, China; Biomedical Intelligent Cloud R&D Center, Xiamen University, Xiamen, China.

Tatiana Agback is with the Department of Molecular Sciences, Swedish University of Agricultural Sciences, Uppsala, Sweden.

Vladislav Orekhov is with the Department of Chemistry and Molecular Biology, University of Gothenburg, Gothenburg, Sweden.

applicable to some situations and may compromise the spectra. Thus, a data-driven method is expected.

Recently, deep learning (DL) has been introduced into exponential signal reconstruction and applied in fast biological spectroscopy [23]. It relies on neural networks [24] and massive parallelization with graphic processing units, to greatly reduce the reconstruction time [23]. Other nice examples can be found in the literatures [13, 25-27]. Despite these successful applications of DL in NMR, there are many unsolved but important problems[26]: 1) The lack of robustness and versatility, as the existing network is hard to maintain good reconstruction when mismatches appear between the training and test data. 2) The lack of the network interpretation and guidance to reduce the number of trainable parameters. The above problems greatly limit the widespread usage of DL in this field.

Hence, the question remains, how to design a deep neural network suitable for reconstructing exponential signals, and achieve more robust performance and low computational cost?

The effectiveness of merging the optimization and DL methods has been recently successfully demonstrated in artificial intelligence, signal processing, and medical imaging [28-36]. It encourages us to provide a clear guidance for merging the optimization and DL, address the above-mentioned unsolved problems and push further development of exponential signal reconstruction.

In this work, we propose a deep thresholding network, whose architecture is inspired by the classical model-based Compressed Sensing (CS) [15-20], to reconstruct exponential signals. Our design of architecture starts from the sparse prior knowledge and adopts the main idea from the iterative soft-thresholding algorithm (ISTA) [15-20], so we call this sparse Model-inspired Deep thresholding network as MoDern. Notably, MoDern has a compact architecture, which is trained by using solely synthetic data with the exponential functions [13, 23], and can generalize well to experimental undersampled data in various scenarios without re-training. Due to its high robustness in respect to mismatch of the training and test data, such as NUS densities and spectra sizes, we further develop an online, open-access, high-performance, and easy-to-use cloud computing platform for MoDern, called XCloud-MoDern, to serve more researchers in biological spectroscopy.

Our main contributions are summarized as follows:

1) The architecture of MoDern follows the backbone of the iterative sparse reconstruction. We design a learnable soft-thresholding to adaptively eliminate the spectrum artifacts introduced by undersampling.

2) We explain the network by analyzing the intermediate reconstruction and the corresponding reconstruction error of peak amplitudes.

3) Extensive experiments on both synthetic and biological NMR data show the merits of MoDern on significantly improving the reconstruction performance under the mismatch between the training and test data.

The remainder of this paper is organized as follows. Section II introduces backgrounds. Section III and IV describes the proposed method and implementations. Section V shows the experiment results. Section VI and VII provide the discussions and conclusion.

## II. BACKGROUND

### A. Exponential Modeling in General Signal Processing

Exponential function is a fundamental signal form in signal processing. In many applications [1-10], signal can be approximated modeled as the sum of exponentials:

$$r_n = \sum_{j=1}^{J}(a_j e^{i\phi_j})e^{-\frac{n\Delta t}{\tau_j}}e^{i\Delta t 2\pi f_j n}, \ n=1,2,...,N, \quad (2)$$

where $i$ is the imaginary unit, $n$ is the index of data points, $\Delta t$ is the time interval, $J$ is the number of spectral peaks, $a_j, f_j, \tau_j, \phi_j$ are the amplitude, frequency, decay time, and phase of the $j^{th}$ spectral peak, respectively. Thus, the sampled data can be represented as a vector $\mathbf{r}=[r_1,r_2,...,r_N]^T \in \mathbb{C}^N$. By performing Fourier transform $\mathbf{F}$ on the signal $\mathbf{r}$, a corresponding spectrum $\mathbf{x}=\mathbf{Fr} \in \mathbb{C}^N$ is obtained.

### B. Fast Biological Spectroscopy

Multi-dimensional biological NMR spectroscopy serves an indispensable and widely used biological tool in modern biology [37, 38], chemistry [39], and life science [40], but suffers from the long data acquisition time. And the duration of an NMR experiment increases rapidly with the dimensionality and spectral resolution. For example, conventional multi-dimensional NMR takes two to six weeks of measurements per structure [11]. Given the importance and time bottleneck of biological spectroscopy, fast sampling and reliable reconstruction is highly expected, called fast biological spectroscopy in this paper.

The time-domain signal in biological NMR which acquired from spectrometer, called free induction decay (FID), is commonly modeled using exponential form [8, 22, 41] in (2). NUS is commonly used to accelerate the data acquisition by sampling partial FIDs. Following the principle of multi-dimensional NMR, the NUS is performed on the $Z$-1 dimension of a $Z$-dimensional NMR since the data acquisition of the last dimension (also called the direct dimension) is not time consuming [22, 42]. Fig. 1 shows the NUS for 2D ($Z$=2) and 3D ($Z$=3) NMR. Only the one dimension ($t_1$) and the one plane ($t_1$-$t_2$) are partially acquired for 2D and 3D NMR, respectively.

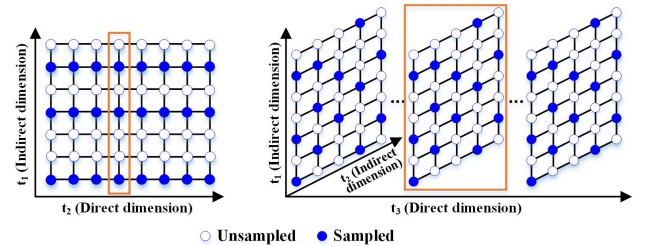

Fig. 1. The sketch map of NUS in (a) 2D and (b) 3D NMR fast spectroscopy. The region marked by an orange rectangle represents one reconstruction.

*C. Sparse Spectra Reconstruction*

CS is one of the state-of-the-art model-based methods for exponential signal reconstruction [15-20]. It exploits the spectral sparsity and the iterative algorithm to reconstruct the undersampled data.

Assuming the sparsity of the spectrum $\mathbf{x}$, the reconstruction problem of CS can be modeled as [17, 18, 43]

$$\min_{\mathbf{x}} \frac{1}{2}\|\mathbf{y} - \mathbf{UF}^H \mathbf{x}\|_2^2 + \lambda \|\mathbf{x}\|_1, \quad (3)$$

where $\mathbf{y} \in \mathbb{C}^M$ is the time-domain signal undersampled by matrix $\mathbf{U} \in \mathbb{R}^{M \times N}$, $\mathbf{F}^H$ is the inverse Fourier transform, $\lambda$ is the regularization parameter that balances the data consistency and the sparsity. $\|\cdot\|_1$ and $\|\cdot\|_2$ denote the $l_1$ and $l_2$ norm, respectively. ISTA is a classical algorithm to solve (3), and its $k^{th}$ iteration process can be written as follows [17, 18, 43, 44]:

$$\begin{cases} \mathbf{d}_k = \mathbf{x}_{k-1} + \mathbf{FU}^T \left( \mathbf{y} - \mathbf{UF}^H \mathbf{x}_{k-1} \right) \\ \mathbf{x}_k = \mathcal{S}(\mathbf{d}_k; \theta) \end{cases}, \quad (4)$$

where $\mathcal{S}(x;\theta) = \max\{|x|-\theta, 0\} \cdot x/|x|$ is the soft-thresholding operator, $\theta$ is the threshold, and $k = 1, 2, ..., K$ is the number of iterations. Initialized with a spectrum $\mathbf{x}_0 = \mathbf{FU}^T \mathbf{y}$ with strong artifacts, CS will reconstruct "clean" spectrum by iteratively alternating the two steps in (4): data consistency and soft-thresholding.

Nevertheless, CS still has some limitations: 1) The manual-setting sparseness prior knowledge makes it suboptimal for reconstructing weak and broad peaks [8] (See Section V.B); 2) With the decrease of the NUS densities and the increase of the spectra size, the computational time for spectra reconstruction increases significantly. For instance, it takes hundreds of seconds to reconstruct a 3D data (See Section V.C). The above issues drive development of new techniques.

*D. Deep Learning Spectra Reconstruction*

Previously, our group proposed a data-driven method DLNMR [23] to reconstruct exponential signals and has aroused great interest in fast biological spectroscopy [26, 45]. It is based on a convolutional neural network DenseNet [46] and refines the intermediately reconstructed spectra by enforcing data consistency. Spectrum artifacts in the frequency domain, which are introduced by NUS data, can be gradually removed in the network. As a proof of concept, the network training can be achieved using solely synthetic data, and then applied to biological NMR data.

However, DLNMR still have some problems, such as seriously compromised performance under the mismatch between the training and test data, and distortions (or disappearance) of weak spectral peaks (See Section V.B and V.C). Thus, further improvement is expected.

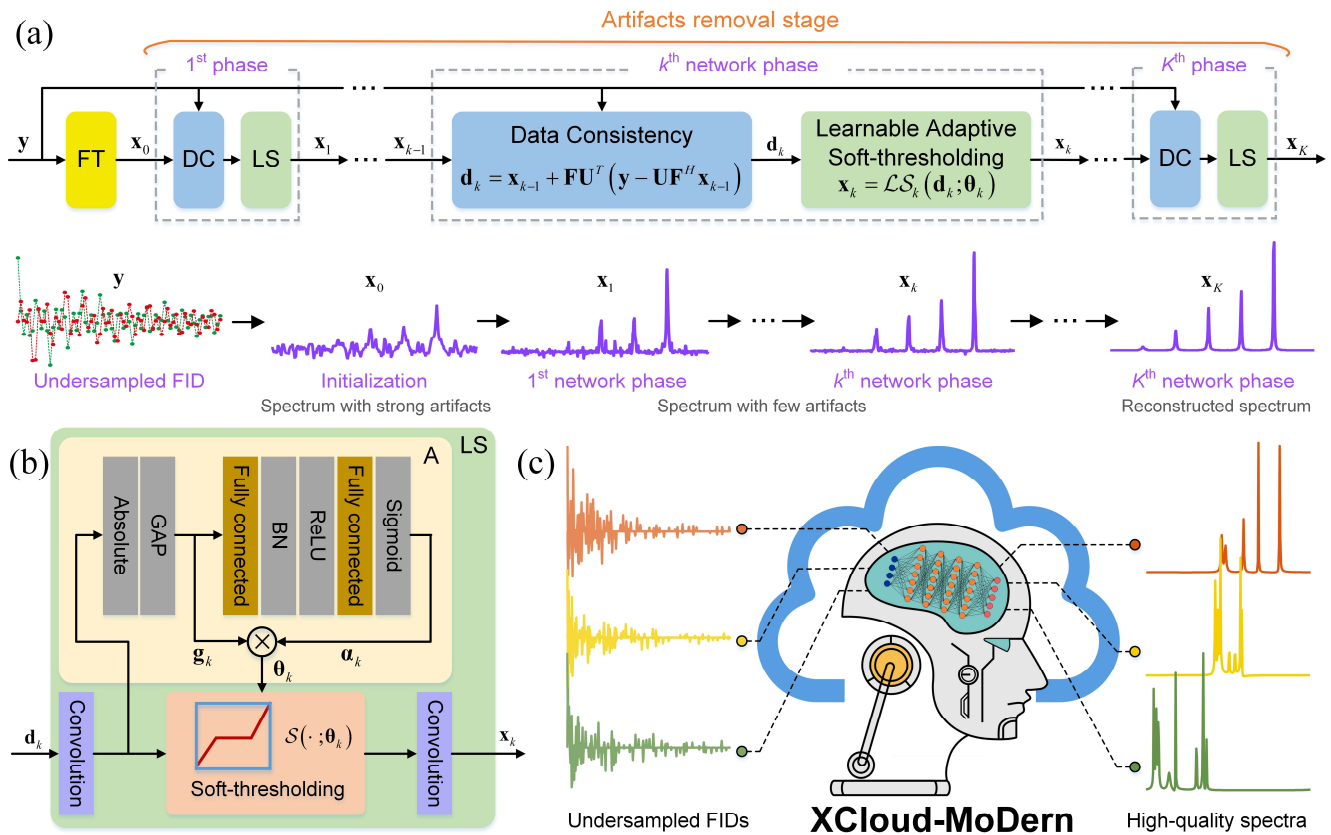

Fig. 2. An overview of MoDern and XCloud-MoDern. (a) The recursive MoDern framework that alternates between the data consistency (DC) module and the learnable adaptive soft-thresholding (LS) module. With the increase of the network phase, artifacts are gradually removed, and finally a high-quality reconstructed spectrum can be obtained. (b) The detailed structure of the learnable adaptive soft-thresholding (LS) module and threshold adaptive-setting (A) module. (c) The developed artificial intelligence cloud computing platform (XCloud-MoDern) for processing multi-dimensional NMR spectra. Note: "FT" is the Fourier transform.

## III. Proposed Method

As shown in Fig. 2(a), once the overall number of iterations is fixed, the backbone of CS can be viewed as our unfolded deep network MoDern. Each network phase of MoDern consists two modules: the data consistency module and the learnable adaptive soft-thresholding module, which correspond to the first and second step of (4), respectively. The overall number of network phases is $K$.

### A. Data Consistency Module

In this module, each spectrum is forced to maintain the data consistency to the sampled signal, which can ensure reconstructed spectra are aligned to acquired data. Exactly same to CS, i.e., the first part of (3), the data consistency module is designed as

$$\mathbf{d}_k = \mathbf{x}_{k-1} + \mathbf{FU}^T\left(\mathbf{y} - \mathbf{UF}^H \mathbf{x}_{k-1}\right), \quad (5)$$

where $\mathbf{d}_k$ is the output of the data consistency module of $k^{th}$ network phase. To show the function of this module more intuitively, (5) is equivalent to

$$\left(\mathbf{F}^H \mathbf{d}_k\right)_n = \begin{cases} \mathbf{F}^H \mathbf{x}_{k-1}, & n \notin \Omega \\ \mathbf{y}, & n \in \Omega \end{cases}, \quad (6)$$

where $n$ is the index of time-domain signals and $\Omega$ is the set of the sampled positions in time-domain signals. And (6) implies that, at the sampled positions, the points should be replaced with the original inputs $\mathbf{y}$, while the update of the unsampled points depends entirely on the reconstruction results of the network. In our implementation, the initial spectrum that inputs the neural network is $\mathbf{x}_0 = \mathbf{FU}^T \mathbf{y}$. The initial input $\mathbf{x}_0$ is with strong artifacts since those unsampled signals are filled with zeros on non-acquired positions.

### B. Learnable Adaptive Soft-thresholding Module

In this part, the motivation for proposing the learnable adaptive soft-thresholding module is introduced, and the architectures of the two designed sub-networks, i.e., basic network and enhanced adaptive one (MoDern), are showed detailedly.

#### 1) Theoretical Background and Motivation

Soft-thresholding operation is always used to enforce the sparsity in the signal processing [31, 32, 44, 47, 48]. In conventional iterative methods [15-20], the optimal value of the threshold is always manually chosen so that all the undersampling artifacts are removed while most of the meaningful peaks are preserved well. However, the proper setting of thresholds is still a challenging issue. The integration of soft-thresholding and deep network opens a promising way to remove artifact-related features, and the proper thresholds can be automatically selected using a back-propagation gradient descent algorithm.

Moreover, soft-thresholding is particularly suitable for deep neural networks in the task of noise and artifact removal due to following two advantages:

1) Similar to the commonly used ReLU, the derivation of soft-thresholding is either one or zero (Fig. 3(b)), which has low complexity and can effectively prevent gradient exploding and vanishing.

2) Different from setting all negative features to zero in ReLU, soft-thresholding sets the near-zero information to zeros and preserves useful negative features (Fig. 3(a)). Thus, it pushes the mean unit activations closer to zero than ReLU in our network (0.18 versus 0.52 in Fig. 4), which can speed up learning by bringing the normal gradient closer to the unit natural gradient [49].

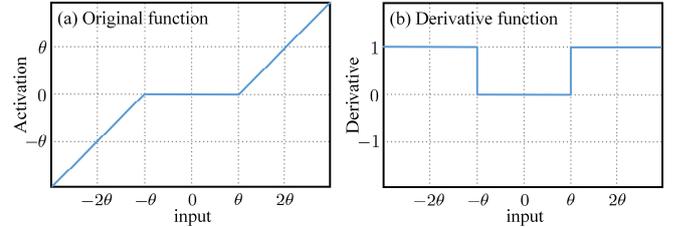

Fig. 3. Soft-thresholding. (a) Original function. (b) derivation function. Note: $\theta$ is the threshold.

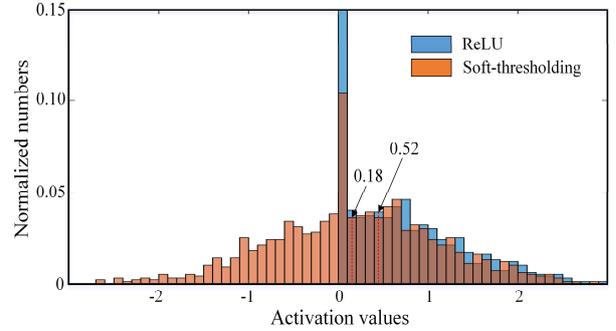

Fig. 4. Distribution of the mean of activation of ReLU and soft-thresholding in our network (use exponential signals as inputs). The distribution of soft-thresholding ($\theta$=0.1) is close to zero than ReLU, i.e., 0.18 versus 0.52.

Here, we first proposed a basic soft-thresholding sub-network architecture. Then in order to further improve the reconstruction performance and generalization, we specifically design the learnable adaptive one. The two strategies of thresholds determination are introduced as follows.

#### 2) Basic Soft-thresholding Architecture

Inspired by CS, we retain the soft-thresholding operator to remove artifacts, and discard a lot of redundant convolution layers which are often used in existing data-driven DL methods [13, 23, 25-27]. It makes the network architecture easier to understand and reduces the computational complexity dramatically. Fig. 6(b) shows the basic soft-thresholding architecture, which consists of a soft-thresholding operator and two convolution layers, and can be represented as

$$\mathbf{x}_k = \mathcal{BS}_k\left(\mathbf{d}_k; \boldsymbol{\theta}_k\right), \quad (7)$$

where $\mathbf{x}_k$ is the output of this module of $k^{th}$ network phase, $\mathcal{BS}$ represents the basic soft-thresholding operator, the threshold $\boldsymbol{\theta}_k$ are used in the soft-thresholding operator $\mathcal{S}$ which is mentioned above in (4).

Notably, in this strategy, after network training, the thresholds are fixed and cannot change with the characteristics of the input, leading to a decline in network generalization.

*3) Adaptive Soft-thresholding Architecture*

To further improve the reconstruction performance and generalization, we elaborately devise the enhanced learnable adaptive soft-thresholding module, which leverages the frequency characteristics difference across distinct features to adaptively rescale weightings of the network. Its key advantage is defining optimal thresholds for each network phase, so that input signals have their own sets of proper thresholds to eliminate the artifact-related features.

For instance (Fig. 5), with the increase of the NUS density of the input signals (i.e., the reduction of initial artifacts), the initially learnt threshold decreases. Moreover, with the increase of network phases, artifacts are gradually removed and thresholds become smaller, so that the reconstruction of strong peaks to weak peaks can be realized successively (See Fig. 8 in Section III.C and Supplement S6).

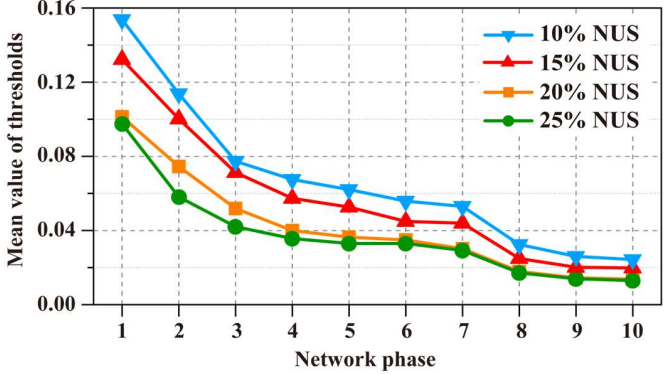

Fig. 5. The variation of adaptive-setting thresholds when signals with different NUS densities are input in a network trained at 15% NUS density. Note: The mean values of thresholds are computed over all channels in the corresponding learnable adaptive soft-thresholding module.

As shown in Fig. 2(b), this module is composed of a soft-thresholding operator, two convolution layers, an absolute operator, a global average pooling (GAP), two fully-connected (FC) layers, a batch normalization (BN), and non-linear functions (ReLU and Sigmoid). Notably, GAP is an operation that calculates a mean value from each channel of feature maps [50]. It can reduce the number of weights to be used in the FC layers, and the features learned by the network are not affected by the changing of artifact locations.

The specific process in $k^{th}$ network phase is described as follows:

1) Threshold adaptive-setting. First, we use one convolution layer for $\mathbf{d}_k$ to realize feature extraction, then GAP is applied to the absolute values of the feature maps to encode the features of the entire space on each channel as global features $\mathbf{g}_k$. After that, $\mathbf{g}_k$ is fed into a two-layer fully-connected network which contains Sigmoid at the end, so that the excitation value $\mathbf{\alpha}_k \in (0,1)$ can be obtained. The adaptive selected threshold is

$$\mathbf{\theta}_k = \mathcal{A}_k(\mathbf{d}_k) = \mathbf{\alpha}_k \times \mathbf{g}_k, \qquad (8)$$

where $\mathcal{A}$ represents the threshold adaptive-setting operator. $\mathbf{\alpha}_k = [\alpha_{1,k}, \alpha_{2,k}, ..., \alpha_{C,k}]^T \in \mathbb{R}^C$ is the multi-channel excitation value, $\mathbf{\theta}_k = [\theta_{1,k}, \theta_{2,k}, ..., \theta_{C,k}]^T \in \mathbb{R}^C$ is the multi-channel threshold, the threshold of $c^{th}$ channel is obtained as follows:

$$\theta_{c,k} = \alpha_{c,k} \times average(|d_{w,h,c}|), \qquad (9)$$

where $w, h, c$ are the index of width, height, and channel of the feature map $d$, respectively. This arrangement is motivated by the fact the threshold not only needs to be positive, but also cannot be too large, thereby preventing the output features from being all zeros.

2) Soft-thresholding. The threshold $\mathbf{\theta}_k$ are used in the soft-thresholding operator $\mathcal{S}$ which is mentioned above in (4). Notably, an individual threshold is applied to each channel of the feature map. After that, we use one convolution layer to combine the multi-channel results after the soft-thresholding and restore the number of channels to the same as the input of the module.

Thus, the learnable adaptive soft-thresholding module is designed as

$$\mathbf{x}_k = \mathcal{LS}_k(\mathbf{d}_k; \mathbf{\theta}_k), \qquad (10)$$

where $\mathbf{x}_k$ is the output of the learnable adaptive soft-thresholding module of $k^{th}$ network phase, $\mathcal{LS}$ represents the adaptive soft-thresholding operator.

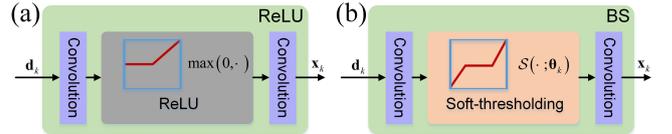

Fig. 6. The detailed structure of the module with (a) ReLU and (b) basic soft-thresholding (BS). Note: For comparison with the module finally adopted in MoDern, the learnable adaptive soft-thresholding (LS) module in Fig. 2 is replaced by (a) and (b), respectively, while other modules remain unchanged.

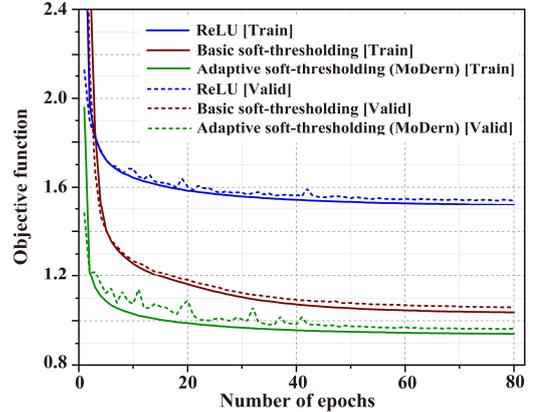

Fig. 7. The objective function among three strategies. Note: Solid and dashed lines indicate an objective function of the training and validation, respectively. [Train]: training stage, [Valid]: validation stage.

To validate the importance of the learnable adaptive soft-thresholding, we show a quantitative comparison of the module with the ReLU (Fig. 6(a)), the basic soft-thresholding (Fig. 6(b)), and the adaptive soft-thresholding (Fig. 2(c)) in the objective function mean square error (MSE). Fig. 7 shows that, in the both training and validation stages, the adaptive one has lower MSE and faster convergence speed, which indicates lower reconstruction error and better generalization, so we finally use this strategy in our MoDern.

In summary, in the proposed MoDern, a data consistency module followed by a learnable adaptive soft-thresholding module constitutes a network phase. As shown in Fig. 2(a), with the increase of network phases, artifacts are gradually removed, and finally a high-quality reconstructed spectrum can be obtained. The pseudo code of MoDern is summarized in Algorithm 1.

---

**Algorithm 1.** Reconstruction with MoDern.

**Input:**
undersampled signal $\mathbf{y}$,
undersampling matrix $\mathbf{U}$,
the overall number of network phases $K$.
**Initialization:**
zero-filled spectrum $\mathbf{x}_0 = \mathbf{F}\mathbf{U}^T\mathbf{y}$,
trained network parameters $\mathbf{\Theta}$.
**Output:**
reconstructed spectrum $\mathbf{x}$.
1: **for** $k = 1 : K$ **do**
2:    Update $\mathbf{d}$ by using (5);
3:    Update $\mathbf{x}$ by using (10);
4: **end for**

---

### C. Network Explainability

To demonstrate the network explainability of MoDern, in Fig. 7, we show the intermediate reconstructed results of a synthetic spectrum with five peaks (See Supplement S1 for parameters, and Supplement S6 for all intermediate reconstruction) and corresponding error of peak amplitudes at each representative network phase.

With the network phase increases, the reconstructed results of MoDern:

1) Visually, the overall lineshapes of the reconstructed spectra gradually approach the fully sampled one, and artifacts near the baseline are eliminated. It means that network pushes the leaked spectrum energy gradually concentrates back to the meaningful peaks, and leads to the accurate spectrum finally (Figs. 8(c)(e)(g)).

2) Quantitatively, the errors of peak amplitudes gradually decrease in the network. Moreover, comparing with the state-of-the-art model-based method (CS [17]) and DL method (DLNMR [23]), the proposed MoDern achieves the lowest reconstruction error (Figs. 8(d)(f)(h)).

In summary, these observations imply that, like the model-based methods, MoDern gradually removes artifacts and reduces the reconstruction error with the increase of network phases. Different from the original end-to-end DL methods, it provides a good explanation of the network reconstruction process. Thus, the proposed MoDern inherits the advantage of both methods.

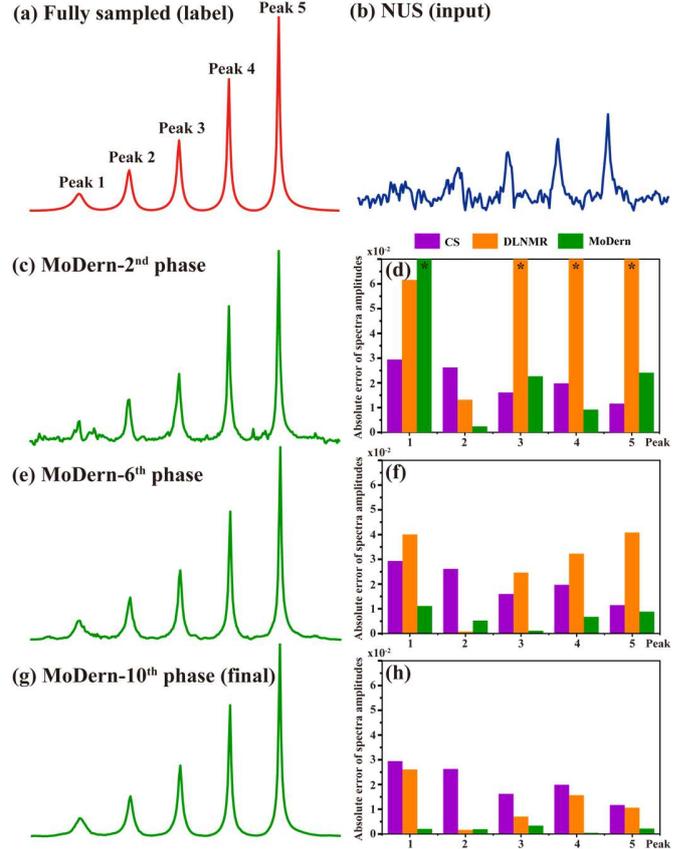

Fig. 8. The reconstructed spectra and absolute error of peak amplitudes at representative network phase. (a) The fully sampled spectrum (label). (b) The undersampled spectrum (input). (c), (e), and (g) are the reconstructed spectra using MoDern at $2^{nd}$, $6^{th}$, $10^{th}$ (final) network phase, respectively. (d), (f), and (h) are the absolute error of peak amplitudes at the corresponding network phase, respectively. Note: "*" means the value is larger than the plotted maximal error. The error of CS in (d), (f), and (h) are the same because it is a model-based method and the final error is chosen.

## IV. IMPLEMENTATION DETAILS

### A. Generation of Training and Validation Datasets

Similar to the previous work [23], the proposed MoDern is trained using solely synthetic data. Here, we generate 40000 fully sampled exponential signals according to (2) and their parameters are in TABLE I. The white Gaussian noise with the standard deviation of $10^{-4}$ is added to each synthetic data $\mathbf{r}_{ref}$ to simulate the real situation. Then 40000 undersampled signals (inputs) $\mathbf{y} = \mathbf{U}\mathbf{r}_{ref}$ and 40000 label spectra $\mathbf{x}_{ref} = \mathbf{F}\mathbf{r}_{ref}$ are generated correspondingly, where $\mathbf{U}$ is the undersampling pattern following the Poisson-gap sampling scheme [51]. Notably, to enrich the diversity of the datasets, with the same NUS density, the undersampling pattern is different for each of 40000 data.

Thus, we obtain 40000 pairs of data $\left(\mathbf{y}^q, \mathbf{U}^q, \mathbf{x}_{ref}^q\right)$, where $q = 1, 2, \ldots, Q$ denotes the $q^{th}$ training sample. Among 40000 pairs of data, 90% (36000 pairs) is for training to obtain the optimal internal parameters and mapping, and 10% (4000 pairs) is for validation to adjust the hyper-parameters and evaluate the reconstruction performance of network preliminarily.

The training and validation datasets are shared at the website: https://github.com/wangziblake/MoDern.

*B. Loss Function and Hyper-parameters*

In the training stage, the loss function is defined as

$$L(\Theta) = \frac{1}{KQ}\sum_{k=1}^{K}\sum_{q=1}^{Q}\left\|\mathbf{x}_{ref}^{q} - \hat{\mathbf{x}}_{k}^{q}\left(\mathbf{y}^{q}, \mathbf{U}^{q}; \Theta_{k}\right)\right\|_{2}^{2}, \quad (11)$$

where $\hat{\mathbf{x}}_k(\mathbf{y}, \mathbf{U}; \Theta_k)$ is the $k^{th}$ output of network phases, $\Theta$ is the set of trainable parameters (TABLE II). The optimal network parameters and mapping is learned by minimizing the (11), and Adam [52] is selected as the optimizer. The overall number of network phases is 10, i.e., $K=10$. He initialization [53] is used to initialize the network weights. The batch size is 10. The initial learning rate was set to 0.001 with an exponential decay of 0.95. The number of epochs is 80, and networks have good convergence on both training and validation datasets. The networks of MoDern for 1D NUS and 2D NUS are trained individually, called as 1D and 2D network respectively.

In the reconstruction stage, for given undersampled exponential signals, we can reconstruct them through the trained MoDern. Notably, MoDern which is trained once can be directly applied to high-quality reconstruction of spectra with different sizes and NUS densities. The robustness verification can be found in Section V.

TABLE I
PARAMETERS OF THE GENERATED SYNTHETIC DATASETS

| Parameters | Minimum | Increment | Maximum |
|---|---|---|---|
| Number of peaks | 1 | 1 | 10 |
| Amplitude | 0.05 | $2.2 \times 10^{-16}$ | 1.00 |
| Frequency | 0.01 | $2.2 \times 10^{-16}$ | 0.99 |
| Decay time | 10.00 | $2.8 \times 10^{-14}$ | 179.20 |
| Phase | 0 | $8.8 \times 10^{-16}$ | $2\pi$ |

Note: The number of peaks increase from 1 to 10 and data points of signals are 256 and 90×44 for 1D and 2D networks, respectively. Each subset contains a specific number of randomly generated peaks and other parameters. Thus, we have 10 subsets and each of them has 4000 data. The increment is obtained by calculating the precision of float numbers.

TABLE II
ARCHITECTURE-RELATED HYPER-PARAMETERS

| Operations | 1D Network for 1D NUS | | 2D Network for 2D NUS | |
|---|---|---|---|---|
| | Hyper-parameters | Output size | Hyper-parameters | Output size |
| Input | / | 1×T×2 | / | 1×($T_1$×$T_2$)×2 |
| 1st Conv | 1×3×2, 32 | 1×T×32 | 3×3×2, 32 | 1×($T_1$×$T_2$)×32 |
| 1st FC | 2 | 1×1×2 | 2 | 1×(1×1)×2 |
| 2nd FC | 32 | 1×1×32 | 32 | 1×(1×1)×32 |
| 2nd Conv (output) | 1×3×32, 2 | 1×T×2 | 3×3×32, 2 | 1×($T_1$×$T_2$)×2 |

Note: "Conv" is the convolution layer, "FC" is the fully-connected layer. Take the 1D network for example, 1, T, 2 in the output size '1×T×2' represent the widths, heights, and channels of the feature map, respectively. 1, 3, 2, 32 in the hyper-parameters of Conv "1×3×2, 32" represent the size and the number of filters. 32 in the hyper-parameters of FC "32" represent the number of neurons.

*C. Extension to the Cloud Computing Platform*

No limited to the local computing, we also implement the proposed MoDern online.

Cloud computing platform is generally web-based and easily all-day accessible through a variety of internet-connected devices without installation. Here, to facilitate the widespread usage of MoDern in data analysis, we develop XCloud-MoDern, an easy-to-use artificial intelligence cloud computing platform, for easily and fast processing the NUS multi-dimensional NMR spectra (Fig. 2(c)). Up to now, the platform supports 2D and 3D spectra online reconstruction using MoDern, and also provides a customized retrospectively undersampling technique (NUS simulator). Take reconstruction tasks as the example, the whole workflow is easy and user-friendly to NMR researchers: 1) Select the dimensionality of the NUS data. 2) Select the configuration of MoDern. 3) Upload the NUS data file and corresponding sampling pattern. 4) Start the reconstruction automatically. 5) Download the reconstructed data file and check the reconstruction time. It is important to mention that, the waiting time is very short due to its high efficiency; data is stored on the cloud for reuse and users can also delete it at any time.

Now, XCloud-MoDern is open access to researchers at http://36.134.147.88:2345/ (Account: CSG-001, Password: CSG@MYTEST_001). Notably, we decide not to open up the registration in the peer review. We have provided manual, demo data and scripts on the cloud for the quick try. The quality and time of reconstruction on cloud (TABLE V) is highly consistent with which on local (TABLE III). We believed that, it will bridge the gap between high-performance and accessible implementations.

## V. EXPERIMENTAL RESULTS

*A. Evaluation Criteria and Compared Methods*

We evaluate the performance of MoDern on both synthetic and biological NMR data through spectral lineshapes and contour maps of reconstructed spectra, peak intensity correlations $R^2$, relative $l_2$ norm error (RLNE) [9], and quantitative measurement.

The square of Pearson linear correlation coefficient $R^2$ is used to measure the correlation of each peak. The relative $l_2$ norm error (RLNE) is used to measure the overall reconstruction error of the spectrum. The $R^2$ and RLNE are defined as

$$R^2(\mathbf{a},\mathbf{b}) = \left(\frac{\sum_{i=1}^{I}(\mathbf{a}_i - \overline{\mathbf{a}})(\mathbf{b}_i - \overline{\mathbf{b}})}{\sqrt{\sum_{i=1}^{I}(\mathbf{a}_i - \overline{\mathbf{a}})^2}\sqrt{\sum_{i=1}^{I}(\mathbf{b}_i - \overline{\mathbf{b}})^2}}\right)^2, \quad (12)$$

$$RLNE(\mathbf{a},\mathbf{b}) = \frac{\|\mathbf{b} - \mathbf{a}\|_2}{\|\mathbf{b}\|_2}, \quad (13)$$

where $\mathbf{a}, \mathbf{b}$ are reconstructed spectra and fully sampled spectra, respectively. $\overline{\mathbf{a}}, \overline{\mathbf{b}}$ are the mean values. The closer the value of $R^2$ gets to 1, the stronger the correlation between the fully sampled spectra and the reconstructed spectra is. The lower RLNE represents the lower reconstruction energy loss (power of RLNE).

The proposed method is mainly compared with the state-of-the-art methods: the model-based CS [17] and the data-driven DLNMR [23]. All local experiments are implemented on a server equipped with dual Intel Xeon CPUs (2.2 GHz, 24 cores in total), 128 GB RAM, and one Nvidia Tesla K40M GPU. The proposed MoDern is performed on Python 3.6 and Tensorflow 1.14.0 [54] as backend, as well as DLNMR. CS is parallelized with aspects to the direct dimensions to maximally reduce the computation time under multiple CPU cores, and two DL methods are trained under one GPU. Specifically, DLNMR owns the same hyper-parameters as the original paper [23], and is trained by training datasets which are same to MoDern. The number of the network parameters of MoDern is about 9% of that needed for DLNMR (5770 versus 54680 and 13450 versus 155000 for 1D and 2D networks, respectively). The total training time of DLNMR are 10.4 hours and 43.7 hours for 1D and 2D networks within 80 epochs, respectively. The total training time of MoDern are 3.9 hours and 15.1 hours for 1D and 2D networks within 80 epochs, respectively.

Moreover, MoDern costs much less reconstruction time with a compact and lightweight architecture (TABLE III). Notably, with the increase of the spectra size and dimensionality, the advantage of MoDern in reconstruction time and memory requirements will become more obvious.

### B. Reconstruction of Synthetic Data

#### 1) Reconstruction on toy examples

First, we use toy examples of a synthetic data with five peaks, to show that the proposed MoDern can overcome the limitation of CS [17] on broad peaks (Fig. 9), and the limitation of DLNMR [23] on the robustness under mismatch between training and test data (Fig. 10).

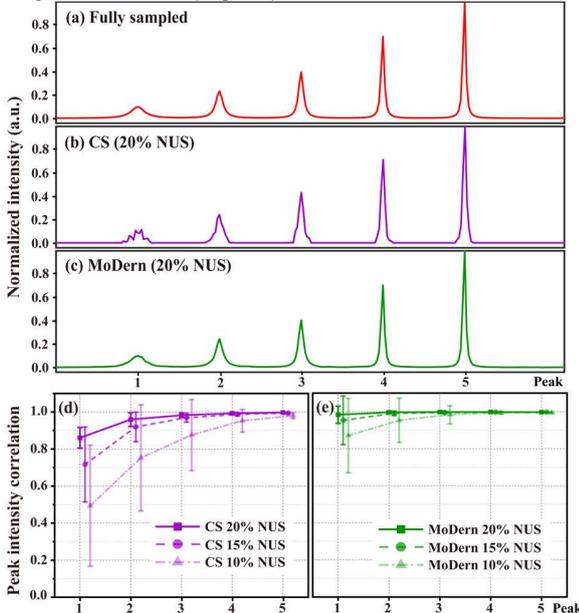

Fig. 9. Reconstruction of a synthetic data with five peaks. (a) The fully sampled spectrum. (b) and (c) are the reconstructed spectra using CS and MoDern from 20% NUS data, respectively. (d) and (e) are $R^2$ between fully sampled spectrum and reconstructed spectrum using CS and MoDern under different NUS densities, respectively. Note: The average and standard deviations of correlations in (d) and (e) are computed over 100 Monte Carlo trials with different sampling masks.

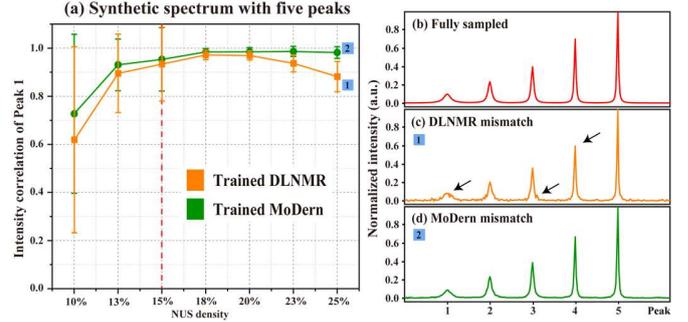

Fig. 10. Reconstruction of the synthetic data under mismatch in DL. (a) The correlation $R^2$ of the lowest Peak 1. (b)-(d) are the fully sampled spectrum, the typical reconstruction by using DLNMR and MoDern, respectively. Note: Both DL networks are trained at NUS density of 15% (the dashed red line in (a)), while the test density varies from 10% to 25%. The average and standard deviations of correlations in (a) are computed over 100 Monte Carlo trials with different sampling masks.

Figs. 9(a)-(c) implies that MoDern provides faithful reconstruction of all five peaks but CS causes visible distortions of broad and weak Peak 1 and 2. This conclusion is further verified in Figs. 9(d)-(e), when the NUS density is at the range 10%~20%, MoDern obtains much higher and more robust peak intensity correlations on Peak 1 and 2.

Fig. 10 shows that, MoDern can maintain the high-quality reconstruction performance even the NUS density of the spectrum is far from trained one (Fig. 10(a)), and lineshapes of the spectra are close to the fully sampled spectra (Fig. 10(d)). However, DLNMR leads to significantly declined performance, such as much lower correlation $R^2$ of Peak 1 (Fig. 10(a)), peak intensity distortions and artifacts (marked arrows in Fig. 10(c)).

#### 2) Reconstruction under different scenarios

Here, we further evaluate the reconstruction performance on synthetic data under different scenarios, such as different number of exponentials (spectra peaks) and NUS densities. Notably, here, we further added DHMF [13], a state-of-the-art DL method for exponential signal reconstruction, as the comparison.

The signals are generated following (2) with parameters listed in TABLE I, and are not included in training datasets. For each (*NUS density, number of exponentials*) pair, 100 Monte Carlo trails are conducted. We set two error thresholds (RLNE=0.05 and 0.02), which are represented by white and red lines, respectively. The corresponding energy loss is 0.25% and 0.04%, which can be considered as small and very small reconstruction error [9, 13].

Fig. 11 shows that MoDern can better handle exponential signal reconstructions with large number of exponentials or low NUS density. The region below the red line, denoting a lower reconstruction error (RLNE) than 0.02, is significantly larger for MoDern than three compared methods. If we relax the acceptable energy loss to 0.25% (white line), the regions of three compared methods obviously become large, whereas MoDern consistently owns the largest region. For example, when NUS density is 15%, MoDern enables the reliable reconstruction of 9 exponentials, while CS, DLNMR, and DHMF allow 2, 4, and 7 exponentials, respectively. These observations imply that MoDern provides better reconstructions than compared methods.

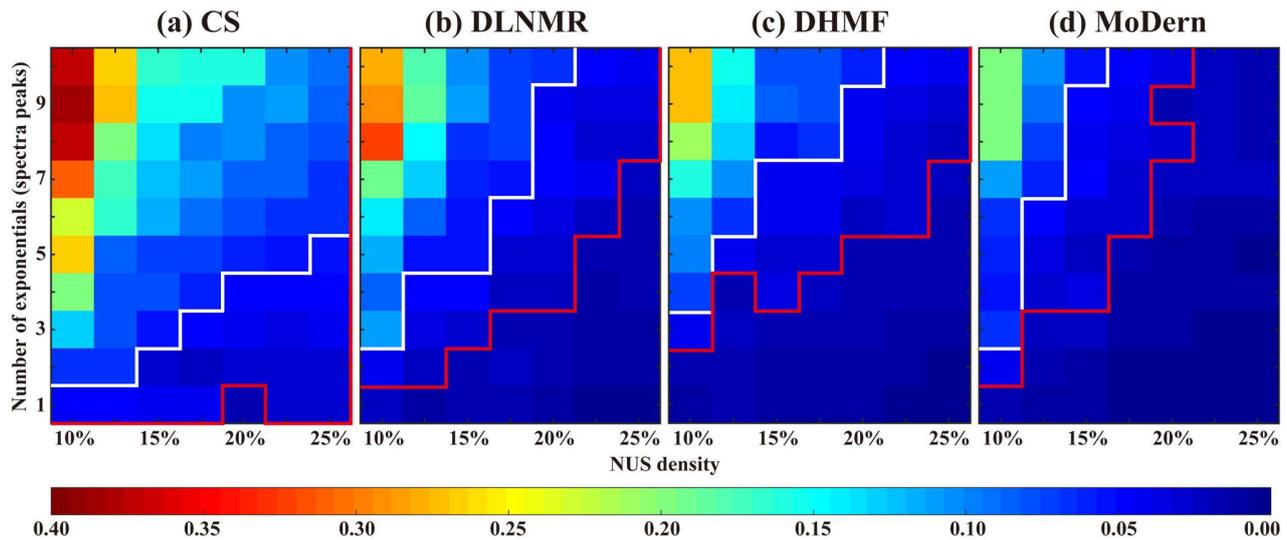

Fig. 11. Reconstruction of the synthetic data under different scenarios. (a)-(d) are average reconstruction errors, RLNEs of CS, DLNMR, DHMF and MoDern, respectively. Note: Each color reflects the average RLNEs over 100 Monte Carlo trials with different sampling masks. Red (or white) line indicates an empirical boundary where the error threshold RLNE is 0.02 (or 0.05). Below the boundary, the reconstruction error of the region is less than the error threshold.

## C. Reconstruction of Biological NMR Data

### 1) Biological NMR data acquisition

The information for biological NMR data used in this paper, including several representative 2D and 3D spectra of proteins and the metabolite mixture, are listed in TABLE III. Details about experimental setups can be found in Supplement S1. All NMR spectra are processed using NMRPipe [55].

TABLE III
INFORMATION FOR BIOLOGICAL NMR DATA AND RECONSTRUCTION TIME (UNITS: SECONDS)

| Spectra type | | Spectra sample | Spectra size | Reconstruction time on local | | |
|---|---|---|---|---|---|---|
| | | | | CS | DLNMR | MoDern |
| 2D | NOESY | Ubiquitin | 928×512 (Full) | 2.39 | 0.85 | **0.17** |
| | HSQC | CD79b | 116×256 (Full) | 0.59 | 0.08 | **0.04** |
| | HSQC | Gb1 | 1146×170 (Full) | 3.32 | 0.62 | **0.13** |
| | TROSY | Ubiquitin | 512×128 (Full) | 1.82 | 0.17 | **0.05** |
| 3D | HNCACB | GB1-HttNTQ7 | 879×90×44 (Full) | 225.12 | 13.08 | **3.18** |
| | HNCO | MALT1 | 735×57×70 (30% NUS) | 135.19 | 11.88 | **2.75** |
| | HNCO | Alpha-synuclein | 221×64×64 (15% NUS) | 16.53 | 3.68 | **0.78** |

Note: For 2D (3D) spectra, the size of the directly detected dimension is followed by the size(s) of the indirect dimension(s).

For fully sampled data from spectrometer, the full spectra are served as the golden standard in validation. The FID were retrospectively undersampled according to Poisson-gap sampling [51]. For NUS sampled data from spectrometers, we also performed further undersampling on the partial FID to emulate lower NUS densities.

In biological NMR, the unit of chemical shift is usually expressed in part per million (ppm) instead of the common Hz, to avoid ambiguity when spectrometers are in different magnet strengths. The chemical shift is defined as

$$chemical\ shift\ (ppm) = \frac{f_{test} - f_{ref}}{f_{spec}} \times 10^6, \qquad (14)$$

where $f_{test}$ is resonance frequency of the sample, $f_{ref}$ is the absolute resonance frequency of a standard compound measured in the same magnetic field, $f_{spec}$ is the frequency of magnetic field strength of the spectrometer.

### 2) Reconstruction under mismatch

A key factor that limits wide usage of existing DL for NMR spectra reconstruction is the lack of robustness and versatility. They cannot overcome the mismatch between training and test data in practical applications. It means that NMR researchers need to spend numerous times on re-training networks to handle various reconstruction tasks of different NUS densities, which is obviously unacceptable.

Here, we evaluate the reconstruction performance under mismatch on two NMR data (a 2D $^1$H-$^{15}$N HSQC spectrum of CD79b and a 3D HNCACB spectrum of GB1-HttNTQ7). To focus on the mismatch problem in DL, the CS method is not included here to save pages but provided in Supplement S2.

In Fig. 12, MoDern and DLNMR are trained using 15% (or 10%) NUS density dataset, to reconstruct 2D (or 3D) NMR data under different NUS densities. MoDern maintains the high-quality and significantly better than DLNMR reconstruction performance when the NUS density of spectra significantly deviates from the level used in the training (Figs. 12(a)(e)).

We want to point out that the observed phenomenon of MoDern is highly aligned to model-based methods and follows our intuition: The higher NUS densities, the better reconstruction qualities. On the contrary, with more given sampled points if the mismatch exists, DLNMR may even show obvious intensity distortions, artifacts, and lower $R^2$ (Fig. 12(c)(g)). More results on other NMR data can be found in Supplement S3.

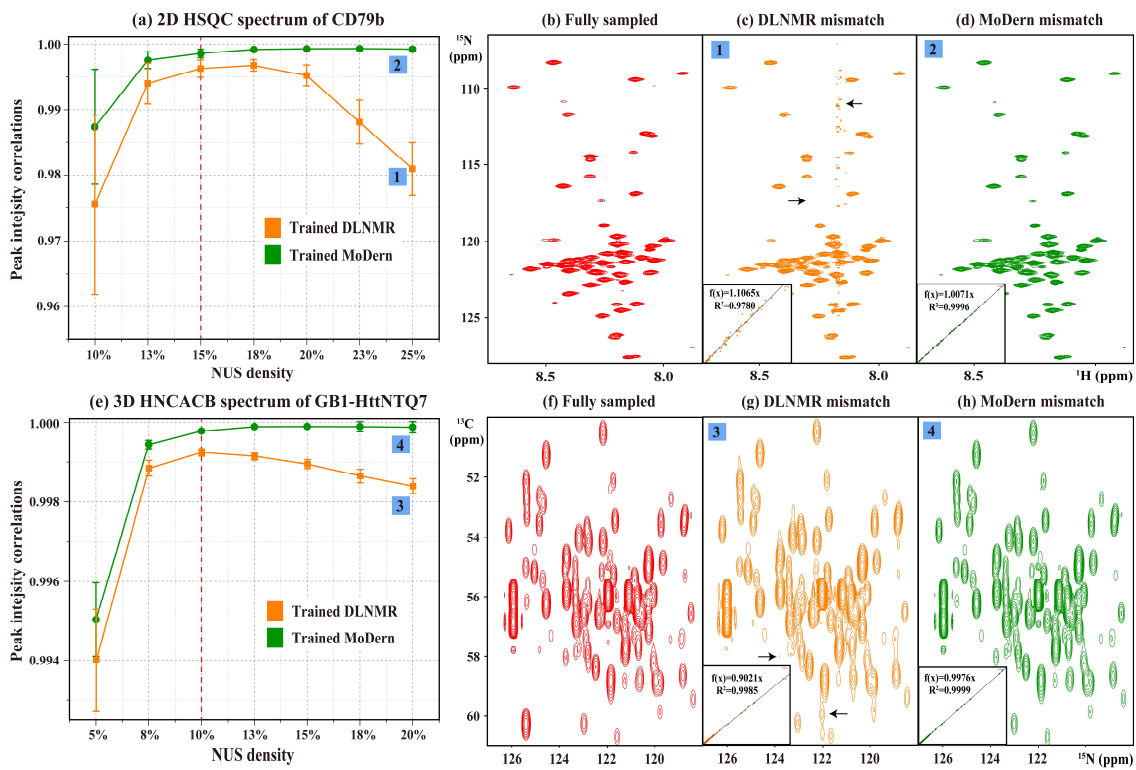

Fig. 12. Reconstruction of 2D and 3D NMR data under mismatch in DL. (a) $R^2$ between the fully sampled 2D HSQC spectrum of CD79b and reconstructed spectra. (b)-(d) are the fully sampled spectrum, the typical reconstructions by DLNMR and MoDern, respectively. (e) $R^2$ between the fully sampled 3D HNCACB spectrum of GB1-HttNTQ7 and reconstructed spectra. (f)-(h) are sub-regions of the projections on $^{13}C$-$^{15}N$ planes of the fully sampled spectrum, the typical reconstructions by DLNMR and MoDern, respectively. Note: The dashed red lines in (a) and (e) indicate the NUS densities that the networks are trained for. The insets of (c)(d)(g)(h) show $R^2$. The average and standard deviations of correlations in (a) and (e) are computed over 100 and 50 Monte Carlo trials with different sampling masks, respectively. The obvious intensity distortions and artifacts are marked with the black arrow.

Moreover, the strong applicability of MoDern to the spectra size can be demonstrated through reconstructions of multiple sets in the above-mentioned TABLE III and the following TABLE IV.

Thus, MoDern is proposed as a robust and versatile technique that can work effectively in a wide range of scenarios without further training, since introduction of optimization idea endows it a certain degree of stability guarantees and the advantages of the adaptive soft-thresholding strategy which is mentioned above.

*3) Reconstruction of high dynamic range data*

High dynamic range means that the difference between the intensity of highest and lowest peaks is in 1~3 orders, raising the challenging reconstruction of weak peaks.

To further demonstrate the reliability of MoDern, here, we perform reconstructions on a high dynamic range 2D $^1H$-$^1H$ NOESY spectrum of Ubiquitin with many weak peaks (the highest intensity of the spectral peak is 600 times of the lowest intensity one).

In Fig. 13(d), we observe that MoDern obtains $R^2$=0.9931 for the intensities of the weak peaks (at the range $0\sim3\times10^6$, i.e., 4% of the highest peak intensity), which outperforms CS ($R^2$=0.9821) and DLNMR ($R^2$=0.9708). Besides, both CS and DLNMR obviously lose some weak peaks marked with the black arrows (Figs. 13(b)(c)). These observations imply that MoDern provides the most faithful reconstruction of low-intensity peaks in the challenging NMR experiments.

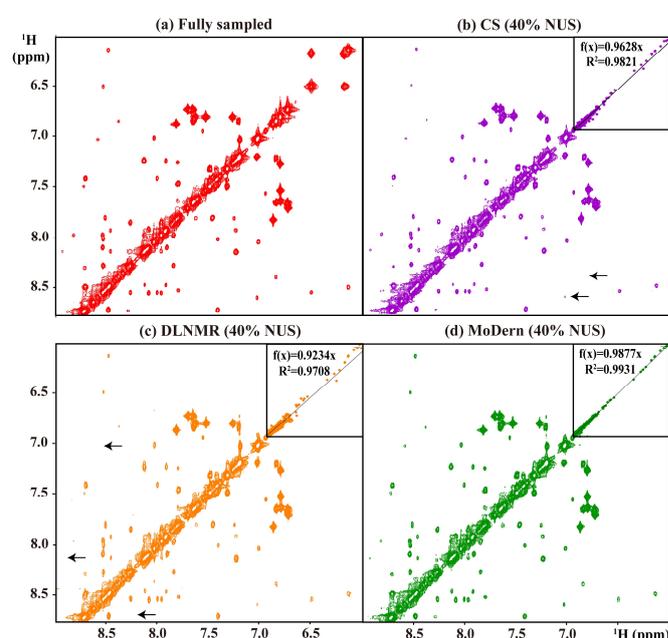

Fig. 13. Reconstruction of a high dynamic range 2D $^1H$-$^1H$ NOESY spectrum of Ubiquitin. (a) The fully sampled spectrum. (b)-(d) are the reconstructed spectra using CS, DLNMR, and MoDern from 40% NUS data, respectively. Note: The insets of (b)-(d) show $R^2$ between fully sampled spectrum and reconstructed spectrum for weak peaks. The obvious peak losses are marked with the black arrow.

TABLE IV
$R^2$ COMPARISON OF THREE METHODS RECONSTRUCTIONS OF 2D BIOLOGICAL NMR DATA UNDER DIFFERENT NUS DENSITIES

| Spectra type, sample, size | Method | $R^2$ under different NUS densities | | | |
|---|---|---|---|---|---|
| | | 10% | 15% | 20% | 25% |
| HSQC, CD79b, 116×256 | CS | 0.9097 ± 0.0239 | 0.9799 ± 0.0088 | 0.9970 ± 0.0019 | 0.9992 ± 0.0003 |
| | DLNMR | 0.9870 ± 0.0052 | 0.9963 ± 0.0013 | 0.9986 ± 0.0005 | 0.9995 ± 0.0001 |
| | MoDern | **0.9934 ± 0.0038** | **0.9986 ± 0.0006** | **0.9995 ± 0.0001** | **0.9998 ± 0.0001** |
| HSQC, Gb1, 1146×170 | CS | 0.9223 ± 0.0239 | 0.9887 ± 0.0083 | 0.9986 ± 0.0018 | 0.9998 ± 0.0001 |
| | DLNMR | 0.9898 ± 0.0038 | 0.9969 ± 0.0006 | 0.9970 ± 0.0005 | 0.9968 ± 0.0004 |
| | MoDern | **0.9942 ± 0.0025** | **0.9986 ± 0.0004** | **0.9994 ± 0.0002** | **0.9999 ± 0.0001** |
| TROSY, Ubiquitin, 512×128 | CS | 0.6635 ± 0.0659 | 0.8660 ± 0.0408 | 0.9659 ± 0.0246 | 0.9880 ± 0.0076 |
| | DLNMR | 0.8945 ± 0.0438 | 0.9604 ± 0.0084 | 0.9738 ± 0.0055 | 0.9795 ± 0.0028 |
| | MoDern | **0.9032 ± 0.0372** | **0.9775 ± 0.0088** | **0.9856 ± 0.0038** | **0.9906 ± 0.0021** |

Note: The average and standard deviations of correlations are computed over 100 Monte Carlo trials with different sampling masks, respectively.

Moreover, TABLE IV demonstrates that, for other three 2D spectra with moderate dynamic range, MoDern can also obtain the higher peak intensity correlations $R^2$ and more robust reconstruction fidelity than compared CS and DLNMR methods under different NUS densities from 10% to 25%. (The typical reconstructed spectra can be seen in Supplement S4).

*4) Quantitative measurement on the relative concentration*

Quantitative measurement on the relative concentration is analyzed on a mixture of three metabolites, including D-Glucose, β-Alanine and Valine. Details about experimental setups can be found in Supplement S1. A time-zero $^1$H-$^{13}$C HSQC spectrum (HSQC$_0$) have better quantitative analysis characteristics that the signal intensities are proportional to concentrations of individual metabolites, and can be obtained by extrapolating a series of 2D HSQC$_i$ spectra ($i=1,2,3$) to zero time through a linear regression extrapolation [56]:

$$\ln(A_{i,n}) = \ln(A_{0,n}) + i \times \ln(F_{A,n}), \quad i = 1,2,3, \quad (15)$$

where $A_{i,n}$ is the peak volume (the integrated signal intensity) of the $n^{th}$ peak in HSQC$_i$, and $A_{0,n}$ is the peak volume of the $n^{th}$ peak in time-zero HSQC$_0$. Notably, the $F_{A,n}$ is free of attenuation during the coherence transfer period, and $A_{0,n}$ is the amplitude attenuation factor for the $n^{th}$ peak. The $A_{0,n}$ is the criteria to measure the relative concentration of metabolites.

Fig. 14(a) shows the spectral peak attribution of three metabolites. Furthermore, to improve the relative concentration of the mixture, we average the peak intensity of individual metabolites, and the relative concentration of each metabolite is calculated as the volume ratio of a substance to Valine in Fig 14(b). More detailed about the concentrations of a mixture can be found in Supplement S5.

Fig. 14(b) shows that MoDern provides the closest relative concentration to that of the fully sampled spectrum than CS and DLNMR using 20% NUS data. Thus, MoDern may be applicable for quantitative measures in high-throughput metabolomics experiments [57].

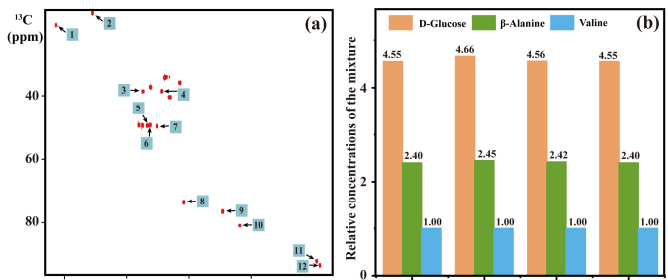

Fig. 14. Quantitative measurement on the relative concentration. (a) Spectral peak attribution for a mixture of three metabolites. Here, twelve cross-peaks are labelled. Among them, Peaks 1-6 are assigned to D-Glucose, Peaks 8-9 are assigned to β-Alanine, and Peaks 7, and 10-12 are assigned to Valine. (b) Relative concentrations of the metabolite mixture. From left to right are the fully sampled spectrum, reconstructed results using CS, DLNMR, and MoDern from 20% data, respectively.

## VI. DISCUSSIONS

### A. Reconstruction of More 3D Biological NMR Data

For ultra-fast high-quality reconstruction of 3D NMR spectra of small, large, and intrinsically disordered proteins, MoDern also shows its great performance. In Fig. 15(a), for 3D HNCACB spectrum of a small protein GB1-HttNTQ7, the peak intensity correlation $R^2$ reaches 0.99 even at 20 times acceleration, i.e., 5% NUS, which can reduce the NMR experimental time from about 22 hours to 1 hour. Figs. 15(b)(c) illustrate that, for a large protein MALT1 and an intrinsically disordered protein Alpha-synuclein, MoDern also provides high-quality spectra reconstruction using only 10% NUS data, while DLNMR appears some artifacts (Supplement S4). Thus, MoDern allows a huge time saving for the experiment of these multi-dimensional NMR data.

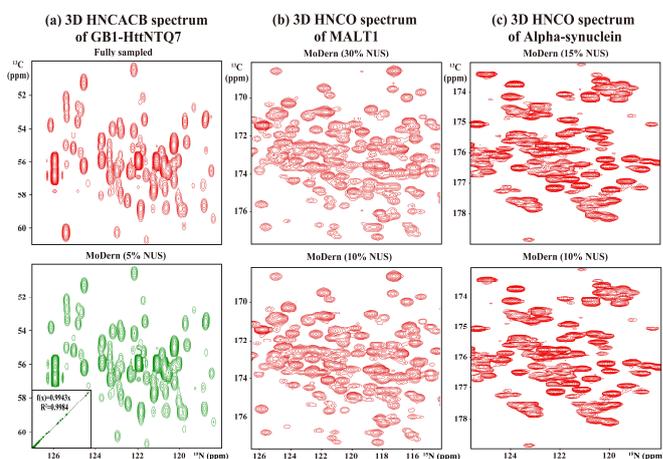

Fig. 15. Reconstruction of three 3D spectrum of the small, large, and intrinsically disordered protein. (a) 3D HNCACB spectrum reconstruction for GB1-HttNTQ7. From top to bottom are the sub-regions of $^{13}$C-$^{15}$N projection from the fully sampled spectrum and the reconstructed spectrum using MoDern from 5% data. The inset shows $R^2$. (b) 3D HNCO spectrum reconstruction for a large protein MALT1. From top to bottom are the sub-regions of $^{13}$C-$^{15}$N projection from the reconstructed spectra using MoDern from 30% and 10% data, respectively. (c) 3D HNCO spectrum reconstruction for an intrinsically disordered protein Alpha-synuclein. From top to bottom are the sub-regions of $^{13}$C-$^{15}$N projection from the reconstructed spectra using MoDern from 15% and 10% data, respectively. Note: The biological data for MALT1 and Alpha-synuclein proteins were acquired under 30% and 15% NUS, respectively, so we cannot give the quantitative evaluation. The further retrospectively random undersampling of NUS data is performed to simulate the lower NUS densities.

## B. Online Reconstruction Time of XCloud-MoDern

To ensure the efficiency of our cloud platform XCloud-MoDern, the configuration of the China Mobile e-cloud host is as follows: 8 cores CPU, 64 GB RAM, one Nvidia T4 GPU, and 500 GB SSD. The reconstruction time on cloud using MoDern is reported in TABLE V, which is highly consistent with TABLE III.

TABLE V
ONLINE RECONSTRUCTION TIME (UNITS: SECONDS)

| Spectra type | | Spectra sample | Spectra size | Reconstruction time online |
|---|---|---|---|---|
| | | | | XCloud-MoDern |
| 2D | NOESY | Ubiquitin | 928×512 | **0.55** |
| | HSQC | CD79b | 116×256 | **0.37** |
| | HSQC | Gb1 | 1146×170 | **0.42** |
| | TROSY | Ubiquitin | 512×128 | **0.39** |
| 3D | HNCACB | GB1-HttNTQ7 | 879×90×44 | **2.50** |
| | HNCO | MALT1 | 735×57×70 | **3.09** |
| | HNCO | Alpha-synuclein | 221×64×64 | **0.94** |

## VII. CONCLUSION

In summary, inspired by a link between the model-based compressed sensing and data-driven deep learning, we propose a sparse model-inspired deep thresholding network (MoDern) as a robust, low-computation-cost, high-fidelity, and ultra-fast approach for exponential signal reconstruction. Extensive results of synthetic and biological NMR data demonstrate that it breaks through the performance bottleneck of state-of-the-art methods and achieves astonishing performance improvements. As the proposed network architecture is quite effective and follow the backbone of compressed sensing method, so we believe that this opens an avenue for theoretical analysis to explore its underlying mechanisms.

Moreover, we further develop XCloud-MoDern, an open-access and easy-to-use artificial intelligence cloud computing platform, for fastly processing the NUS multi-dimensional NMR spectra online. We believe that this genre will serve many researchers and benefit numerous NMR applications in biology, chemistry, and life science.

Despite the impressive results of current MoDern, it is noteworthy that the spectra reconstruction from exponential functions is essentially an ill-posed problem [26]. This indicates that, although DL methods leverage a large amount of well-registered data to learn good transformation, it is theoretically impossible for network to realize error-free reconstruction. We are working on characterizing theoretical error bounds for deep neural networks to more effectively guide the design of network architectures and the parameter selections. In the future, we will try more challenging applications, such as higher dimensional data (e.g., 4D and 5D) and other experimental types (e.g., diffusion, dynamic, and relaxation), and extending the method into more fields, such as medical imaging, geoscience, and radar systems. Moreover, for other areas where datasets are relatively scarce, how to mine data characteristics and build synthetic datasets to train deep networks is also a problem worthy of further exploration.

The shared sources about MoDern and XCloud-MoDern can be found at https://github.com/wangziblake/MoDern.

ACKNOWLEDGMENTS

The authors thank Marius Clore and Samuel Kotler for providing the 3D HNCACB data; Jinfa Ying for assisting in processing and helpful discussions on the 3D HNCACB spectrum; Luke Arbogast and Frank Delaglio for providing the 2D HSQC spectrum of Gb1; Chunyan Xiong for helpful discussions on activation functions. The authors thank China Telecom for providing cloud computing service support at initial. The authors thank NMRPipe and MddNMR for sharing 2D and 3D NMR data on the websites; NMRPipe and SPARKY for data processing support. The authors also thank editors and reviewers for the valuable comments.

# Supplementary Material for

# "A Sparse Model-inspired Deep Thresholding Network for Exponential Signal Reconstruction—Application in Fast Biological Spectroscopy"

Zi Wang, Di Guo, Zhangren Tu, Yihui Huang, Yirong Zhou, Jian Wang, Liubin Feng, Donghai Lin, Yongfu You, Tatiana Agback, Vladislav Orekhov, Xiaobo Qu*

## S1. Experimental setups of the synthetic and biological NMR data

### 1.1 1D synthetic data

The parameters of the 1D synthetic spectrum with five peaks are described in Table S1-1.

**Table S1-1. Parameters of 1D synthetic spectrum with five peaks.**

| Parameters | Peak 1 | Peak 2 | Peak 3 | Peak 4 | Peak 5 |
|---|---|---|---|---|---|
| Amplitude | 0.10 | 0.30 | 0.50 | 0.70 | 1.00 |
| Normalized Frequency | 0.17 | 0.33 | 0.50 | 0.67 | 0.84 |
| Decay time | 10.00 | 20.00 | 30.00 | 50.00 | 60.00 |
| Phase | 0 | 0 | 0 | 0 | 0 |

### 1.2 2D biological NMR data

The 2D $^1$H-$^1$H NOESY spectrum of human ubiquitin was acquired from ubiquitin at 298K on an 600 MHz Varian spectrometer as was described previously [1]. The fully sampled spectrum has 928 × 512 complex points, the size of the directly detected dimension ($^1$H) is followed by the size of the indirect dimension ($^1$H).

The 2D $^1$H-$^{15}$N HSQC spectrum of cytosolic CD79b protein was acquired for 300 μM $^{15}$N-$^{13}$C labeled sample of cytosolic CD79b in 20 mM sodium phosphate buffer at 298K on an 800 MHz Bruker spectrometer as was described previously [2, 3]. The fully sampled spectrum has 1024 × 256 complex points, the size of the directly detected dimension ($^1$H) is followed by the size of the indirect dimension ($^{15}$N).

The 2D $^1$H-$^{15}$N HSQC spectrum of Gb1 was acquired from GB1 at 298K on a 600 MHz Bruker spectrometer as was described previously [3]. The fully sampled spectrum has 1676 × 170 complex points, the size of the directly detected dimension ($^1$H) is followed by the size of the indirect dimension ($^{15}$N).

The 2D $^1$H-$^{15}$N TROSY spectrum of ubiquitin was acquired from ubiquitin at 298K on an 800 MHz Bruker spectrometer as was described previously [4]. The fully sampled spectrum has 682 × 128 complex points, the size of the directly detected dimension ($^1$H) is followed by the size of the indirect dimension ($^{15}$N).

The 2D $^1$H-$^{13}$C HSQC spectrum of a mixture of three metabolites, including 24.27 mM D-Glucose, 11.49 mM β-Alanine, 5.38 mM D-Mannose and dissolved in 0.5ml $D_2O$. It was acquired using a phase sequence at 298 K on a Bruker Avance III-HD 850 MHz spectrometer using 5mm CPTCI probe. The fully sampled spectrum has 1024 × 256 complex points, the size of the directly detected dimension ($^1$H) is followed by the size of the indirect dimension ($^{13}$C).

### 1.3 3D biological NMR data

The fully sampled 3D HNCACB spectrum of GB1-HttNTQ7 was acquired at 298K on a 700 MHz Bruker spectrometer as was described previously [5]. The fully sampled spectrum has 1024 × 90 × 44 complex points, the size of the directly detected dimensions ($^1$H) is followed by the size of the indirect dimensions ($^{15}$N and $^{13}$C).

The NUS 3D HNCO spectrum of MALT1 protein was acquired at 298K on a 700 MHz Bruker spectrometer as was described previously [6]. Only 30% NUS data were recorded in the experiment. The expected fully spectrum has 1024 × 57 × 70 complex points, the size of the directly detected dimensions ($^1$H) is followed by the size of the indirect dimensions ($^{15}$N and $^{13}$C).

The NUS 3D HNCO spectrum of Alpha-synuclein protein was acquired at 293K on an 800 MHz Bruker spectrometer as was described previously [7]. Only 15% NUS data were recorded in the experiment. The expected fully spectrum has 1024 × 64 × 64 complex points, the size of the directly detected dimensions ($^1$H) is followed by the size of the indirect dimensions ($^{15}$N and $^{13}$C).

# S2. Comparison with CS on weak peaks reconstruction of NMR data

In Figure S2-1, the results on a 2D $^1$H-$^{15}$N HSQC spectrum of CD79b, are highly consistent with the synthetic data. We first observe qualitatively that the spectrum reconstructed by MoDern is more similar to fully sampled spectrum, which is corroborated by the higher peak intensity correlations than CS. Second, after zooming out 1D $^{15}$N traces to show the lineshapes, we find that several weak peaks are significantly distorted in the CS reconstruction, while it does not occur in the MoDern. And for the general peaks, both methods have good reconstruction.

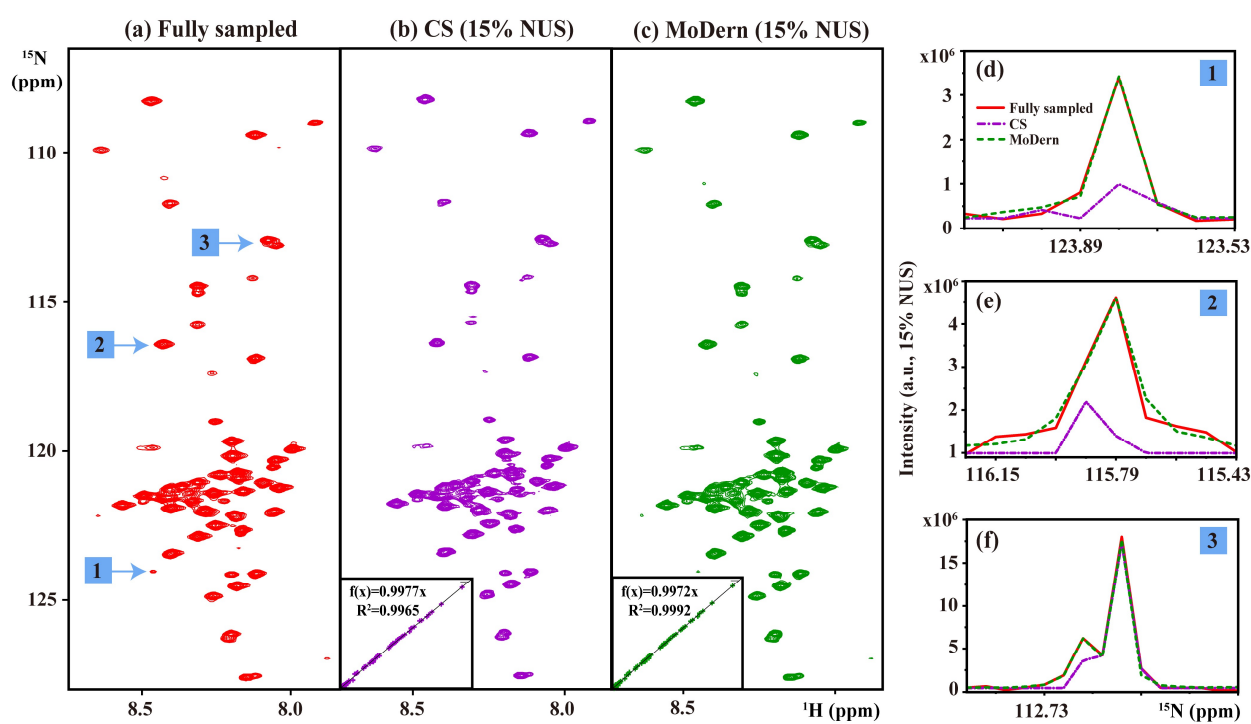

**Figure S2-1.** Reconstruction of 2D $^1$H-$^{15}$N HSQC spectrum of CD79b. (a) The fully sampled spectrum. (b) and (c) are the reconstructed spectra using CS and MoDern from 15% NUS data, respectively. (d-f) are spectral lineshapes. Note: The insets of (b-c) show the peak intensity correlations $R^2$ between fully sampled spectrum and reconstructed spectrum using CS and MoDern, respectively.

# S3. Reconstruction of more other NMR data under mismatch

## 3.1 2D NMR spectra

For the reconstruction of spectra with moderate dynamic range (Figure S3-1 and S3-2), MoDern can maintain the high-quality reconstruction performance even the NUS density of the spectrum is far from trained one, and very high peak intensity correlations (>0.999) can demonstrate this. However, DLNMR shows the significant performance decline, e.g., peak intensity distortions and artifacts, which can be seen reflected in peak intensity correlations and reconstructed spectra.

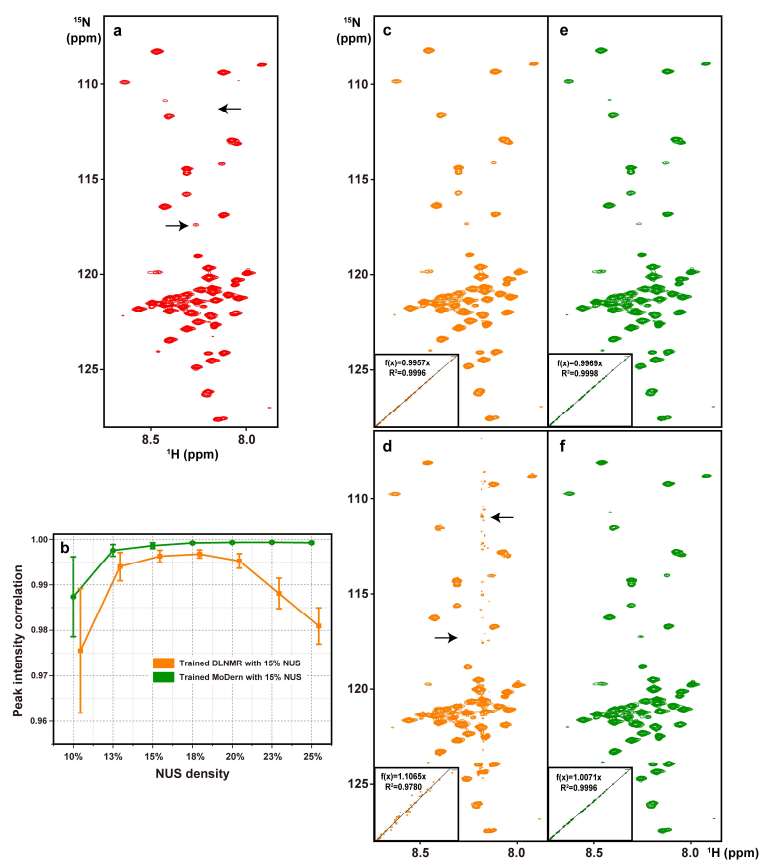

**Figure S3-1.** The reconstruction results of the trained MoDern, and the trained DLNMR from varied NUS density for 2D $^1$H-$^{15}$N HSQC spectrum of CD79b. (a) The fully sampled spectrum. (b) The correlations of DLNMR and MoDern trained using 15% NUS density dataset respectively, to reconstruct spectra sampled under a series of NUS densities ranging from 10% to 25%. (c) and (d) are the typical of reconstructed results of DLNMR trained using 25% and 15% NUS density datasets, respectively, to reconstruct spectra from 25% data. (e) and (f) are the typical of reconstructed results of MoDern trained using 25% and 15% NUS density datasets, respectively, to reconstruct spectra from 25% data. The insets of (c-f) show the peak intensity correlation between fully sampled spectrum and reconstructed spectrum. Note: The average and standard deviations of correlations in (b) are computed over 100 NUS trials. The peak intensity correlations of DLNMR (orange line) are shifted horizontally for clear display but the values unchanged. The intensity distortions and artifacts in (d) are marked with the black arrow.

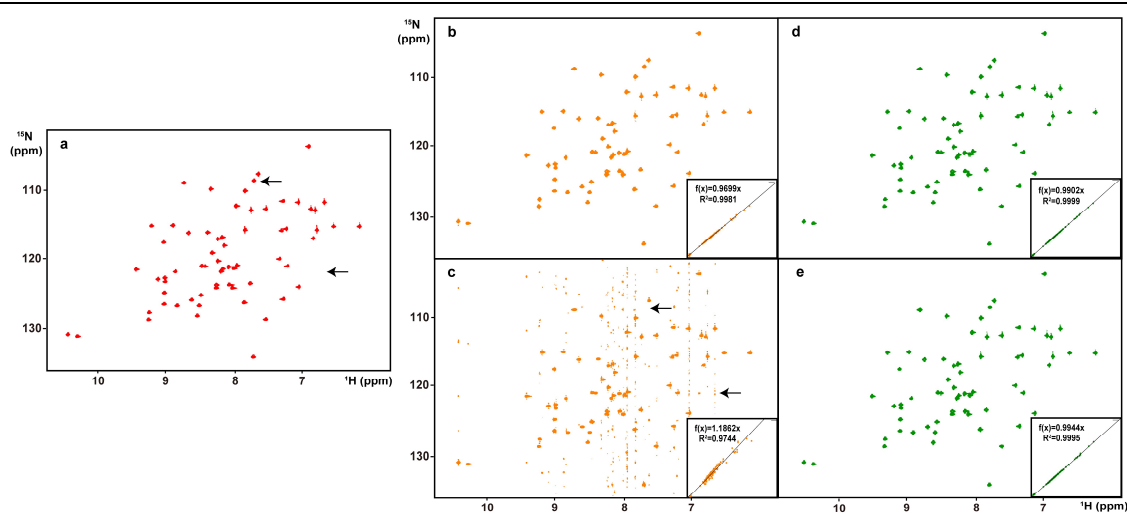

**Figure S3-2.** The reconstruction results of the trained MoDern, and the trained DLNMR from varied NUS density for 2D $^1$H-$^{15}$N HSQC spectrum of Gb1. (a) The fully sampled spectrum. (b) and (c) are the typical of reconstructed results of DLNMR trained using 25% and 15% NUS density datasets, respectively, to reconstruct spectra from 25% data. (d) and (e) are the typical of reconstructed results of MoDern trained using 25% and 15% NUS density datasets, respectively, to reconstruct spectra from 25% data. The insets of (b-e) show the peak intensity correlation between fully sampled spectrum and reconstructed spectrum. Note: The intensity distortions and artifacts in (c) are marked with the black arrow.

## 3.2 3D NMR spectra

In Figure S3-3, MoDern can maintain the high-quality reconstruction performance even the NUS density of the spectrum is far from trained one, and very high peak intensity correlations (>0.996) can demonstrate this. However, DLNMR shows the performance decline, e.g., peak intensity distortions and artifacts, which can be seen reflected in reconstructed spectra.

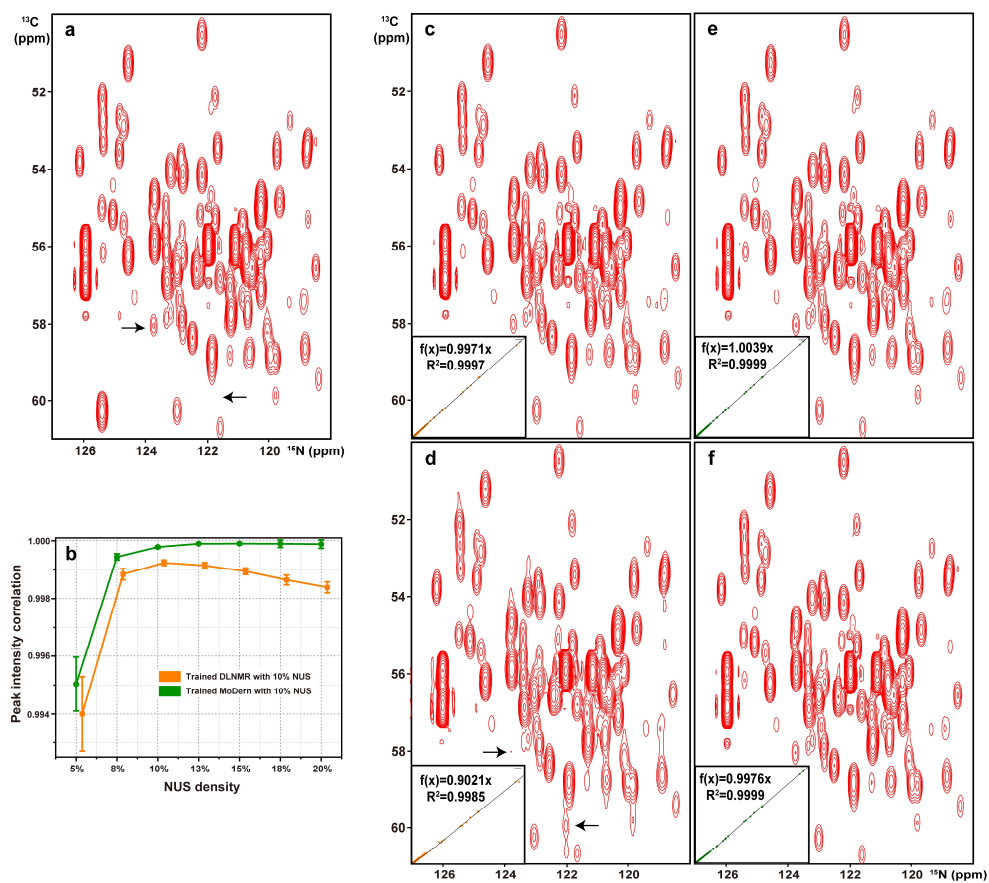

**Figure S3-3.** The reconstructed sub-region of the projections on $^{13}$C-$^{15}$N planes of the trained MoDern, and the trained DLNMR from varied NUS density for 3D HNCACB spectrum of GB1-HttNTQ7. (a) The fully sampled spectrum. (b) The correlations of DLNMR and MoDern trained using 10% NUS density dataset respectively, to reconstruct spectra sampled under a series of NUS densities ranging from 5% to 20%. (c) and (d) are the typical of reconstructed results of DLNMR trained using 20% and 10% NUS density datasets, respectively, to reconstruct spectra from 20% data. (e) and (f) are the typical of reconstructed results of MoDern trained using 20% and 10% NUS density datasets, respectively, to reconstruct spectra from 20% data. The insets of (c-f) show the peak intensity correlation between fully sampled spectrum and reconstructed spectrum. Note: The average and standard deviations of correlations in (b) are computed over 50 NUS trials. The peak intensity correlations of DLNMR (orange line) are shifted horizontally for clear display but the values unchanged. The intensity distortions and artifacts in (d) are marked with the black arrow.

# S4. More details on spectra reconstruction

The proposed MoDern will be compared with two state-of-the-art NMR spectra reconstruction methods, including a model-based optimization method (CS [1]) and a data-driven deep learning method (DLNMR [3]). And all undersampled spectra are generated according to Poisson-gap sampling [8].

## 4.1 Reconstruction of 2D spectra

Fully sampled data were acquired for all 2D spectra. The existence of fully sampled spectra would be helpful serving as the golden standard in reconstruction validation. The undersampled FID were obtained by retrospectively undersampling the fully sampled FID. Here are the results which are not shown in the Main Text.

For the reconstruction of spectra with moderate dynamic range at the NUS density of 20% (Figure S4-1, S4-2), MoDern can obtain very high peak intensity correlation (>0.99), and representative lineshapes of the spectra closing to the fully sampled spectra can demonstrate this. CS and DLNMR can achieve peak intensity correlation, >0.98 and >0.97, respectively.

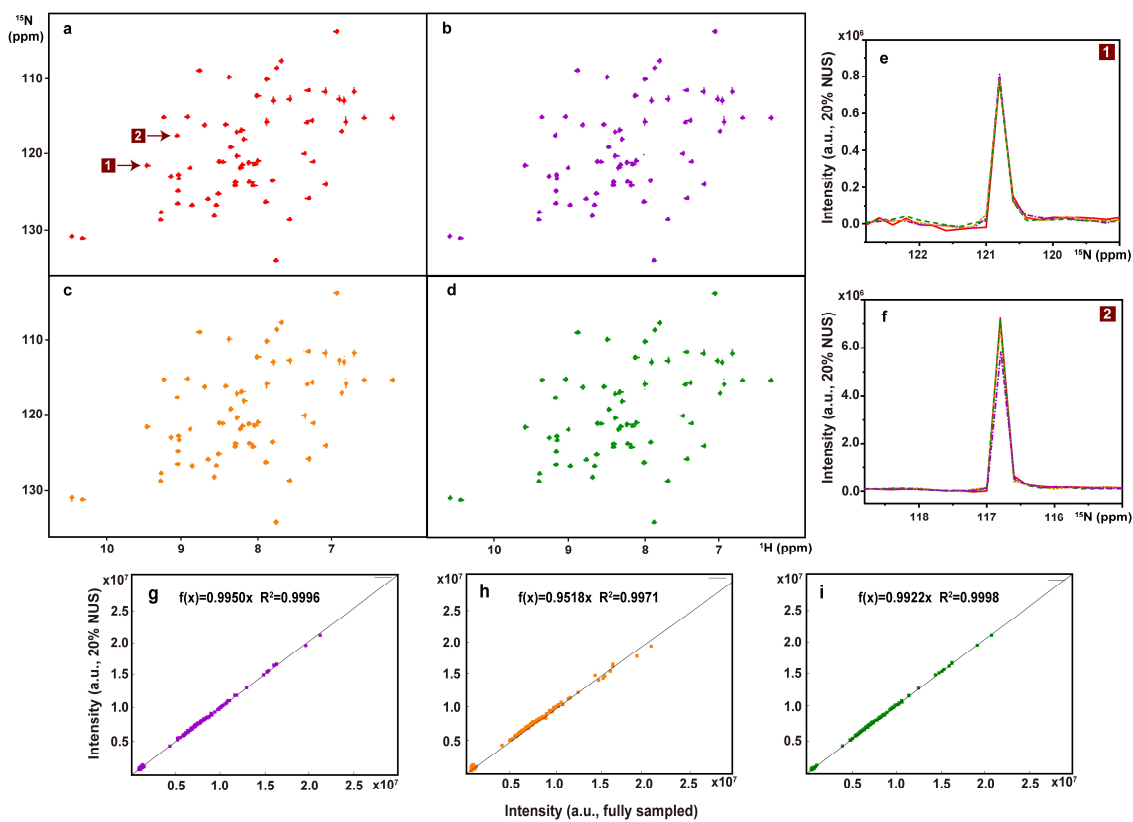

**Figure S4-1.** 2D $^1$H-$^{15}$N HSQC spectrum of Gb1. (a) The fully sampled spectrum. (b), (c) and (d) are reconstructed spectra using CS, DLNMR, and MoDern from 20% data, respectively. (e) and (f) are zoomed out 1D $^{15}$N traces, and the red, purple, orange, and green lines represent the fully sampled spectra, CS, DLNMR, and MoDern reconstructed spectra, respectively. (g), (h), and (i) are the peak intensity correlations obtained by CS, DLNMR, and MoDern, respectively.

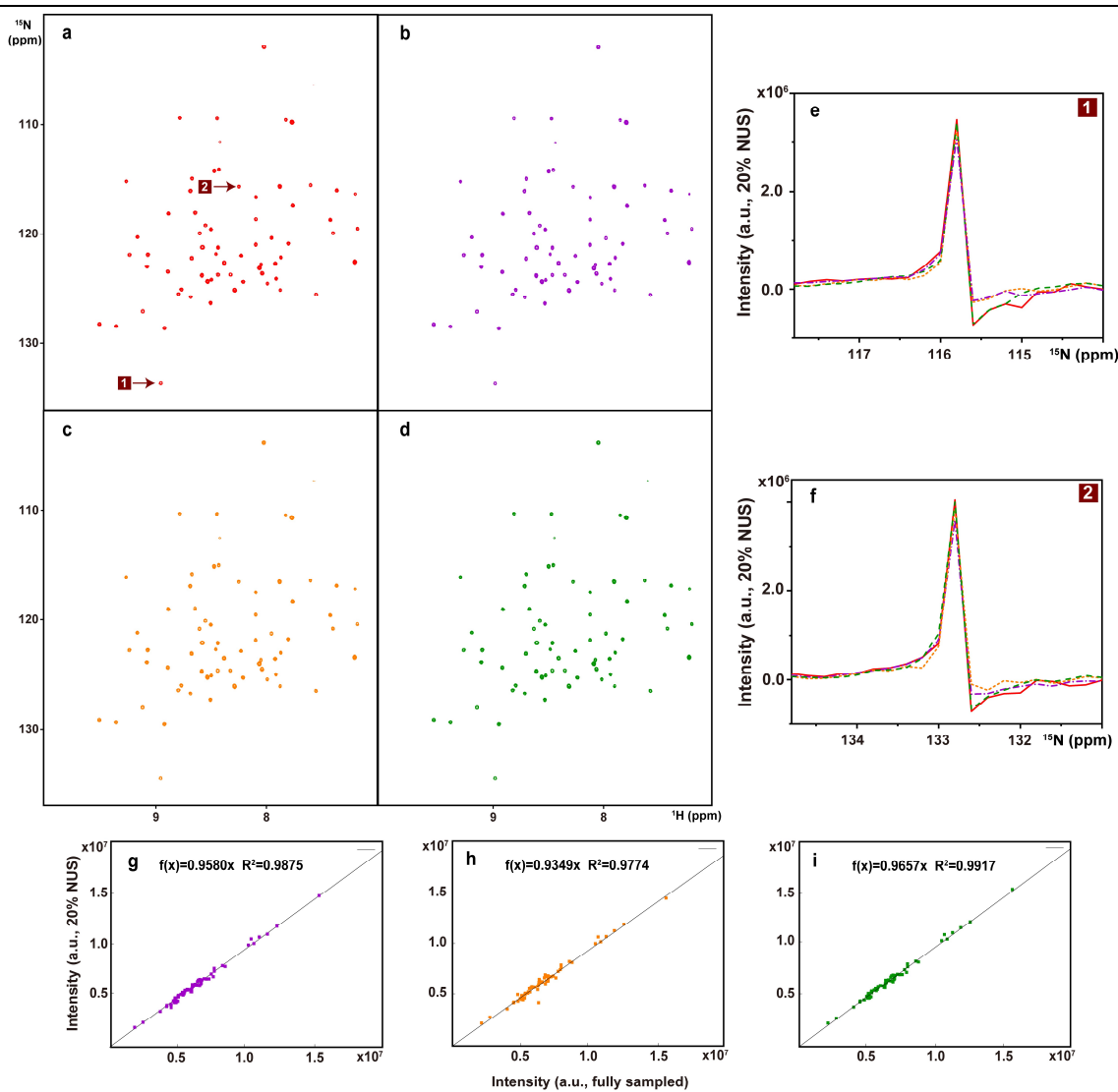

**Figure S4-2.** 2D $^1H$-$^{15}N$ TROSY spectrum of ubiquitin. (a) The fully sampled spectrum. (b), (c) and (d) are reconstructed spectra using CS, DLNMR, and MoDern from 20% data, respectively. (e) and (f) are zoomed out 1D $^{15}N$ traces, and the red, purple, orange, and green lines represent the fully sampled spectra, CS, DLNMR, and MoDern reconstructed spectra, respectively. (g), (h), and (i) are the peak intensity correlations obtained by CS, DLNMR, and MoDern, respectively.

## 4.2 Reconstruction of 3D spectra

### 4.2.1 Reconstruction of 3D spectra of small proteins

Fully sampled FID data were acquired for a small protein GB1-HttNTQ7. The existence of fully sampled spectra would be helpful serving as the golden standard in reconstruction validation. The undersampled FID were obtained by retrospectively undersampling the fully sampled FID.

For the reconstruction of spectra of small proteins (Figure S4-3), all of three methods CS, DLNMR, and MoDern can produces nice reconstructions that are very closing to the fully sampled ones. And the peak intensity correlations of them with $R^2>0.99$ shows the high fidelity of reconstruction.

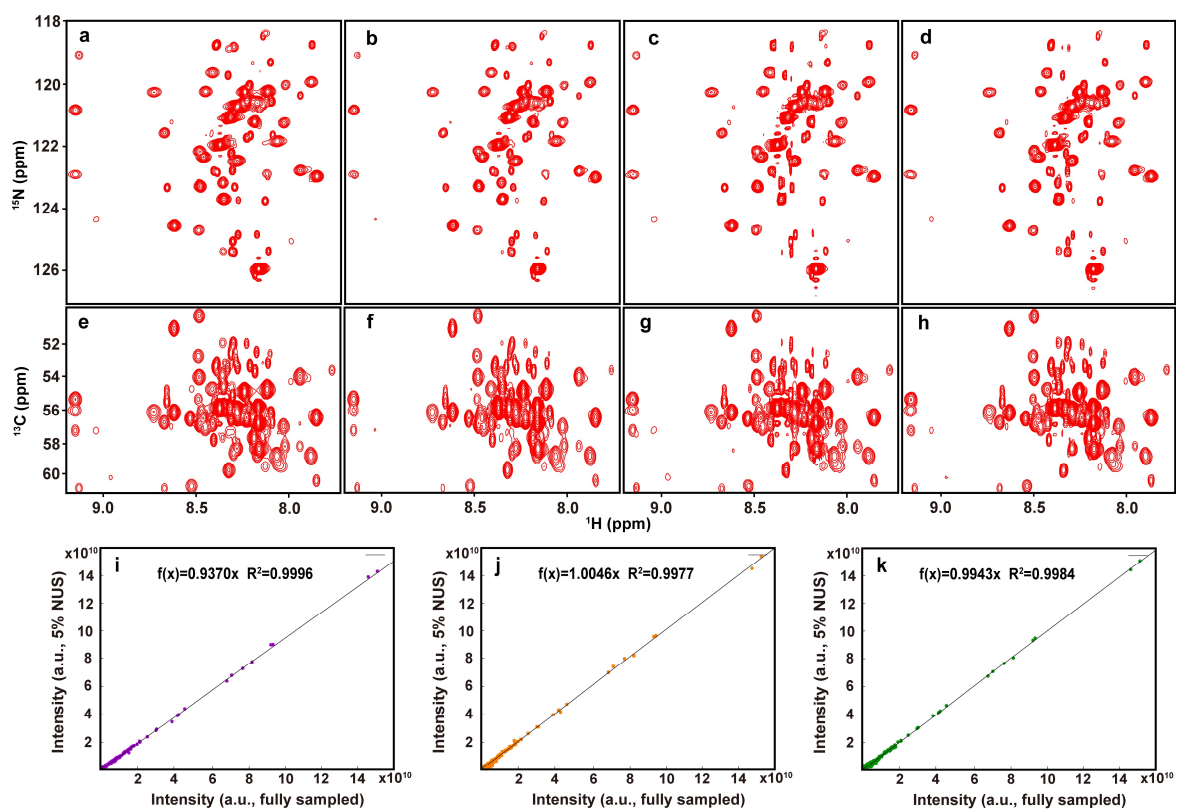

**Figure S4-3.** The sub-region of the projections on $^1$H-$^{15}$N and $^1$H-$^{13}$C planes of the 3D HNCACB spectrum of GB1-HttNTQ7. (a) and (e) are projections of the fully sampled spectrum. (b) and (f) are projections of the CS reconstructed spectrum. (c) and (g) are projections of the DLNMR reconstructed spectrum. (d) and (h) are projections of the MoDern reconstructed spectrum. (i), (j), and (k) are the peak intensity correlations obtained by CS, DLNMR, and MoDern, respectively. Note: 5% NUS data were acquired for reconstruction.

### 4.2.2 Reconstruction of 3D spectra of a large protein and an intrinsically disordered protein

Partial FID data were acquired for a large protein (MALT1) and an intrinsically disordered protein (Alpha-synuclein). These two spectra were experimentally recorded with non-uniform sampling technique in spectrometers for reducing data acquisition time.

For the large protein, i.e., MALT1, the reconstructed spectra were depicted in Figure S2-5. Results show that all of three methods CS, DLNMR, and MoDern can reconstruct the HNCO spectra very well with 30% NUS data, implying that 30% data would be adequate for reconstruction methods to provide reliable results. Here, we performed retrospectively undersampling on the 30% NUS data, taking one out the three data points randomly for emulating the 10% NUS in experiments. The reconstructed spectra by MoDern from 10% NUS data are very similar to the spectra reconstructed using 30% NUS data and the reconstruction performance on relatively weak peaks is better than CS and DLNMR, indicating that MoDern still offers nice reconstruction even under the high acceleration factor.

For an intrinsically disordered protein, i.e., Alpha-synuclein, the reconstructed spectra were shown in Figure S2-6. Results show that all of three methods CS, DLNMR, and MoDern can give nice reconstructions with

15% NUS data. Here, we performed retrospectively undersampling on the 15% NUS data, taking two out the three data points randomly for emulating the 10% NUS in experiments. Even under a higher acceleration, only 10% data used for reconstruction, both MoDern and CS still allow good reconstruction, while DLNMR appears some artifacts.

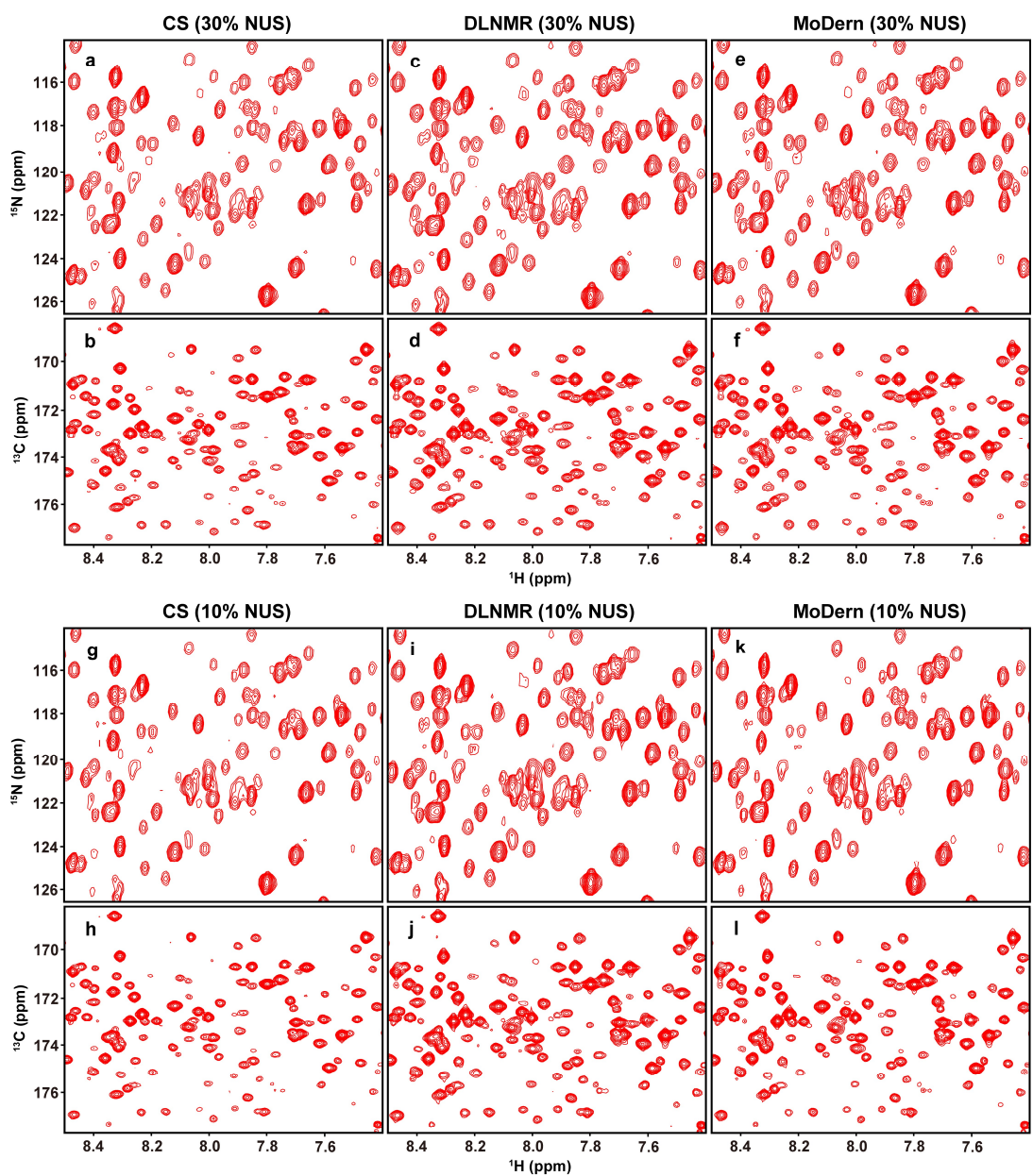

**Figure S4-4.** The sub-region of the projections on $^1$H-$^{15}$N and $^1$H-$^{13}$C planes of the 3D $^1$H-$^{15}$N HNCO spectrum of MALT1 protein. (a-b), (c-d), and (e-f) are projections of the reconstructed spectrum using CS, DLNMR, and MoDern from 30% NUS data, respectively. (g-h), (i-j), and (k-l) are projections of the reconstructed spectrum using CS, DLNMR, and MoDern from 10% NUS data, respectively.

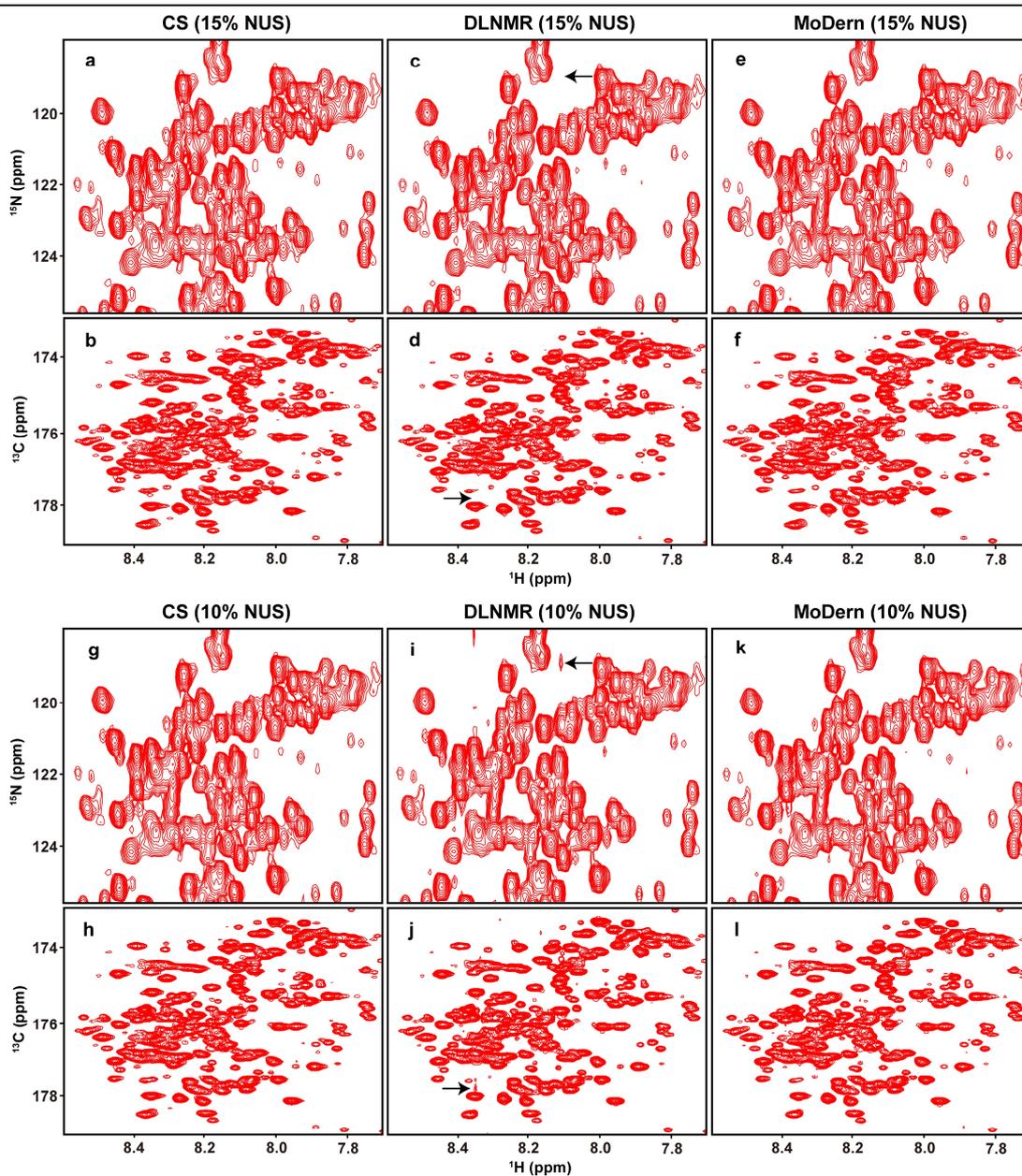

**Figure S4-5.** The sub-region of the projections on $^1$H-$^{15}$N and $^1$H-$^{13}$C planes of the 3D $^1$H-$^{15}$N HNCO spectrum of Alpha-synuclein protein. (a-b), (c-d), and (e-f) are projections of the reconstructed spectrum using CS, DLNMR, and MoDern from 15% NUS data, respectively. (g-h), (i-j), and (k-l) are projections of the reconstructed spectrum using CS, DLNMR, and MoDern from 10% NUS data, respectively. Note: The artifacts in (i-j) are marked with the black arrow.

# S5. Quantitative measurement on concentration of a mixture

Relative concentrations of a mixture (three metabolites, including D-Glucose, β-Alanine and Valine), estimated from fully sampled spectra, and reconstructed spectra using CS, DLNMR, and MoDern from 20% NUS data, respectively, are reported below in Table S5-1, S5-2, S5-3, S5-4.

**Table S5-1.** Extrapolated peak volumes ($A_0$) and measured peak volumes ($A_1$, $A_2$, $A_3$) from fully sampled 2D HSQC$_i$ ($i$ = 1, 2, 3) spectra. Peaks 1, 3, 4, 5 are assigned to D-Glucose (α) while peaks 2 and 6 are assigned to D-Glucose (β). The volume of D-Glucose is the sum of those of D-Glucose (α) and D-Glucose (β). The ratio D-Glucose: β-Alanine: Valine = 6.87: 3.62: 1.51 ≈ 4.55:2.40:1.00.

| Metabolites | | Peak ID | Peak volumes | | | | | |
|---|---|---|---|---|---|---|---|---|
| | | | Measured on spectra | | | Extrapolated from HSQC$_i$ ($i$=1,2,3) | | |
| | | | HSQC$_1$ | HSQC$_2$ | HSQC$_3$ | HSQC$_0$ | Average | Standard deviation |
| D-Glucose | α | 1 | 2.05×10$^9$ | 9.08×10$^8$ | 3.60×10$^8$ | 2.80×10$^9$ | 2.65×10$^9$ | 1.34×10$^8$ |
| | | 3 | 1.90×10$^9$ | 9.33×10$^8$ | 3.76×10$^8$ | 2.59×10$^9$ | | |
| | | 4 | 1.86×10$^9$ | 9.04×10$^8$ | 4.21×10$^8$ | 2.50×10$^9$ | | |
| | | 5 | 2.02×10$^9$ | 9.92×10$^8$ | 4.35×10$^8$ | 2.73×10$^9$ | | |
| | β | 2 | 3.13×10$^9$ | 1.39×10$^9$ | 4.74×10$^8$ | 4.32×10$^9$ | 4.22×10$^9$ | 1.48×10$^8$ |
| | | 6 | 3.14×10$^9$ | 1.55×10$^9$ | 8.89×10$^8$ | 4.11×10$^9$ | | |
| β-Alanine | | 8 | 5.20×10$^9$ | 2.69×10$^9$ | 1.27×10$^9$ | 3.49×10$^9$ | 3.62×10$^9$ | 1.79×10$^8$ |
| | | 9 | 5.53×10$^9$ | 2.71×10$^9$ | 1.18×10$^9$ | 3.75×10$^9$ | | |
| Valine | | 7 | 1.37×10$^9$ | 6.94×10$^8$ | 3.70×10$^8$ | 1.81×10$^9$ | 1.51×10$^9$ | 2.03×10$^8$ |
| | | 10 | 1.05×10$^9$ | 4.97×10$^8$ | 1.93×10$^8$ | 1.44×10$^9$ | | |
| | | 11 | 3.06×10$^9$ | 1.25×10$^9$ | 4.63×10$^8$ | 1.40×10$^9$ | | |
| | | 12 | 3.05×10$^9$ | 1.22×10$^9$ | 4.52×10$^8$ | 1.39×10$^9$ | | |

**Table S5-2.** Extrapolated peak volumes ($A_0$) and measured peak volumes ($A_1$, $A_2$, $A_3$) from reconstructed 2D $HSQC_i$ ($i = 1, 2, 3$) spectra using CS from 20% NUS data. The ratio D-Glucose: β-Alanine: Valine = 6.67: 3.51: 1.43 ≈ 4.66:2.45:1.00.

| Metabolites | | Peak ID | Peak volumes | | | | | |
|---|---|---|---|---|---|---|---|---|
| | | | Measured on spectra | | | Extrapolated from $HSQC_i$ ($i$=1,2,3) | | |
| | | | $HSQC_1$ | $HSQC_2$ | $HSQC_3$ | $HSQC_0$ | Average | Standard deviation |
| D-Glucose | α | 1 | 1.95×10$^9$ | 8.57×10$^8$ | 3.16×10$^8$ | 2.68×10$^9$ | 2.61×10$^9$ | 1.14×10$^8$ |
| | | 3 | 1.95×10$^9$ | 8.76×10$^8$ | 3.43×10$^8$ | 2.66×10$^9$ | | |
| | | 4 | 1.81×10$^9$ | 8.60×10$^8$ | 3.93×10$^8$ | 2.44×10$^9$ | | |
| | | 5 | 1.94×10$^9$ | 9.99×10$^8$ | 3.96×10$^8$ | 2.66×10$^9$ | | |
| | β | 2 | 2.97×10$^9$ | 1.35×10$^9$ | 4.40×10$^8$ | 4.12×10$^9$ | 4.06×10$^9$ | 8.34×10$^7$ |
| | | 6 | 2.99×10$^9$ | 1.75×10$^9$ | 8.57×10$^8$ | 4.00×10$^9$ | | |
| β-Alanine | | 8 | 5.02×10$^9$ | 2.63×10$^9$ | 1.24×10$^9$ | 3.37×10$^9$ | 3.51×10$^9$ | 1.92×10$^8$ |
| | | 9 | 5.38×10$^9$ | 2.64×10$^9$ | 1.15×10$^9$ | 3.64×10$^9$ | | |
| Valine | | 7 | 1.33×10$^9$ | 6.30×10$^8$ | 3.37×10$^8$ | 1.76×10$^9$ | 1.43×10$^9$ | 2.25×10$^8$ |
| | | 10 | 9.18×10$^8$ | 4.19×10$^8$ | 1.69×10$^8$ | 1.25×10$^9$ | | |
| | | 11 | 2.95×10$^9$ | 1.20×10$^9$ | 4.26×10$^8$ | 1.35×10$^9$ | | |
| | | 12 | 2.98×10$^9$ | 1.15×10$^9$ | 4.17×10$^8$ | 1.36×10$^9$ | | |

**Table S5-3.** Extrapolated peak volumes ($A_0$) and measured peak volumes ($A_1$, $A_2$, $A_3$) from reconstructed 2D $HSQC_i$ ($i$ = 1, 2, 3) spectra using DLNMR from 20% NUS data. The ratio D-Glucose: β-Alanine: Valine = 6.94: 3.56: 1.47 ≈ 4.56:2.42:1.00.

| Metabolites | | Peak ID | Peak volumes | | | | | |
|---|---|---|---|---|---|---|---|---|
| | | | Measured on spectra | | | Extrapolated from $HSQC_i$ ($i$=1,2,3) | | |
| | | | $HSQC_1$ | $HSQC_2$ | $HSQC_3$ | $HSQC_0$ | Average | Standard deviation |
| D-Glucose | α | 1 | 1.97×10⁹ | 8.74×10⁸ | 3.28×10⁸ | 2.70×10⁹ | 2.56×10⁹ | 1.17×10⁸ |
| | | 3 | 1.90×10⁹ | 9.31×10⁸ | 3.57×10⁸ | 2.61×10⁹ | | |
| | | 4 | 1.79×10⁹ | 9.19×10⁸ | 4.03×10⁸ | 2.42×10⁹ | | |
| | | 5 | 1.86×10⁹ | 1.02×10⁹ | 4.36×10⁸ | 2.53×10⁹ | | |
| | β | 2 | 3.01×10⁹ | 1.39×10⁹ | 5.06×10⁸ | 4.14×10⁹ | 4.38×10⁹ | 3.44×10⁸ |
| | | 6 | 3.46×10⁹ | 1.75×10⁹ | 8.57×10⁸ | 4.63×10⁹ | | |
| β-Alanine | | 8 | 5.06×10⁹ | 2.64×10⁹ | 1.24×10⁹ | 3.40×10⁹ | 3.56×10⁹ | 2.32×10⁸ |
| | | 9 | 5.51×10⁹ | 2.65×10⁹ | 1.16×10⁹ | 3.73×10⁹ | | |
| Valine | | 7 | 1.35×10⁹ | 6.49×10⁸ | 3.46×10⁸ | 1.79×10⁹ | 1.47×10⁹ | 2.10×10⁸ |
| | | 10 | 1.01×10⁹ | 4.79×10⁸ | 1.88×10⁸ | 1.38×10⁹ | | |
| | | 11 | 2.97×10⁹ | 1.21×10⁹ | 4.48×10⁸ | 1.35×10⁹ | | |
| | | 12 | 2.99×10⁹ | 1.17×10⁹ | 4.29×10⁸ | 1.36×10⁹ | | |

**Table S5-4.** Extrapolated peak volumes ($A_0$) and measured peak volumes ($A_1$, $A_2$, $A_3$) from reconstructed 2D $HSQC_i$ ($i$ = 1, 2, 3) spectra using MoDern from 20% NUS data. The ratio D-Glucose: β-Alanine: Valine = 6.73: 3.55: 1.48 ≈ 4.55:2.40:1.00.

| Metabolites | | Peak ID | Peak volumes | | | | | |
|---|---|---|---|---|---|---|---|---|
| | | | Measured on spectra | | | Extrapolated from $HSQC_i$ ($i$=1,2,3) | | |
| | | | $HSQC_1$ | $HSQC_2$ | $HSQC_3$ | $HSQC_0$ | Average | Standard deviation |
| D-Glucose | α | 1 | $2.02 \times 10^9$ | $8.78 \times 10^8$ | $3.49 \times 10^8$ | $2.75 \times 10^9$ | $2.59 \times 10^9$ | $1.74 \times 10^8$ |
| | | 3 | $1.83 \times 10^9$ | $8.88 \times 10^8$ | $3.56 \times 10^8$ | $2.50 \times 10^9$ | | |
| | | 4 | $1.77 \times 10^9$ | $8.75 \times 10^8$ | $3.94 \times 10^8$ | $2.39 \times 10^9$ | | |
| | | 5 | $2.00 \times 10^9$ | $9.77 \times 10^8$ | $4.16 \times 10^8$ | $2.72 \times 10^9$ | | |
| | β | 2 | $3.05 \times 10^9$ | $1.19 \times 10^9$ | $3.35 \times 10^8$ | $4.24 \times 10^9$ | $4.14 \times 10^9$ | $1.45 \times 10^8$ |
| | | 6 | $3.07 \times 10^9$ | $1.53 \times 10^9$ | $8.52 \times 10^8$ | $4.04 \times 10^9$ | | |
| β-Alanine | | 8 | $5.10 \times 10^9$ | $2.67 \times 10^9$ | $1.26 \times 10^9$ | $3.43 \times 10^9$ | $3.55 \times 10^9$ | $1.72 \times 10^8$ |
| | | 9 | $5.42 \times 10^9$ | $2.65 \times 10^9$ | $1.16 \times 10^9$ | $3.67 \times 10^9$ | | |
| Valine | | 7 | $1.35 \times 10^9$ | $6.87 \times 10^8$ | $3.58 \times 10^8$ | $1.79 \times 10^9$ | $1.48 \times 10^9$ | $2.05 \times 10^8$ |
| | | 10 | $1.00 \times 10^9$ | $4.70 \times 10^8$ | $1.41 \times 10^8$ | $1.40 \times 10^9$ | | |
| | | 11 | $3.02 \times 10^9$ | $1.21 \times 10^9$ | $4.27 \times 10^8$ | $1.38 \times 10^9$ | | |
| | | 12 | $2.99 \times 10^9$ | $1.17 \times 10^9$ | $4.15 \times 10^8$ | $1.37 \times 10^9$ | | |

## S6. The reconstructed spectra at each network phase

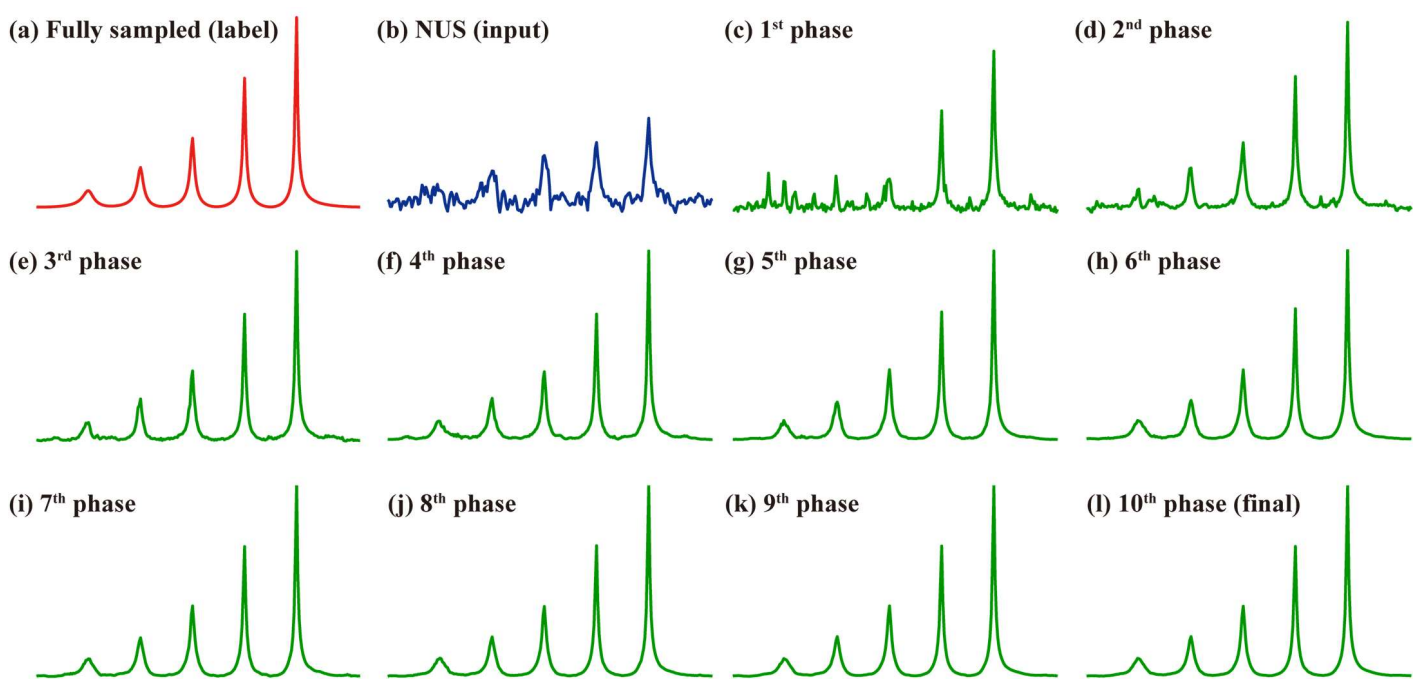

**Figure S6-1.** A typical reconstruction of a synthetic five-peak spectrum. (a) The fully sampled spectrum (label). (b) The undersampled spectrum (input). (c)-(l) are the reconstructed spectra using the proposed MoDern from 1st to 10th network phase, respectively.